\pdfoutput=1

\documentclass[11pt]{article}

\usepackage[final]{acl}

\usepackage{amsmath,amsfonts,bm}

\def\eqref#1{equation~\ref{#1}}

\def\1{\bm{1}}

\DeclareMathAlphabet{\mathsfit}{\encodingdefault}{\sfdefault}{m}{sl}
\SetMathAlphabet{\mathsfit}{bold}{\encodingdefault}{\sfdefault}{bx}{n}

\newcommand{\mytitle}{Dual Debiasing for Noisy In-Context Learning for Text Generation}

\usepackage{microtype}
\usepackage{booktabs} 
\usepackage{bbm}
\usepackage{mathtools}
\usepackage{nccmath}
\usepackage{setspace}

\usepackage{subcaption}
\usepackage{stackengine,scalerel}
\usepackage{accents}
\usepackage{algorithm} 
\usepackage{algorithmic}  
\usepackage[linesnumbered,ruled,vlined,algo2e]{algorithm2e}
\SetKwInput{KwInput}{Input}                
\SetKwInput{KwOutput}{Output}              

\SetCommentSty{mycommfont}

\SetAlCapSty{algcapsty}

\usepackage[T1]{fontenc}
\usepackage{wrapfig,lipsum,booktabs}

\usepackage{soul}
\usepackage{dsfont}
\usepackage{enumerate}
\usepackage{enumitem}

\usepackage{hyperref}
\usepackage{amsmath}
\usepackage{amsfonts}
\usepackage{bbm}
\usepackage{dsfont}
\usepackage[Symbol]{upgreek}
\usepackage{lscape}
\usepackage{caption}
\usepackage{balance}
\usepackage{xspace}
\usepackage{float}
\usepackage{xcolor}
\usepackage{color, colortbl}
\usepackage{wasysym}
\usepackage{multirow}
\usepackage{array, boldline, rotating}

\usepackage{amssymb}
\usepackage{pifont}
\renewcommand*\eqref[1]{(\ref{#1})}

\newcommand{\eg}{\emph{e.g.,~}}
\newcommand{\ie}{\emph{i.e.,~}}

\newcommand{\myparagraph}[1]{\vspace{0.07cm}\noindent\textbf{#1}~}

\def\code#1{\texttt{#1}}

\newcommand{\thickhline}{\hlineB{4}}

\definecolor{LightCyan}{rgb}{0.88,1,1}

\NewDocumentCommand{\supptitle}{s}{
\onecolumn
\begin{center}
    \rule{\textwidth}{0.03cm}\\[0.1cm]
    -Supplementary Material-\\[0.2cm]
    {\Large 
        \textbf{\mytitle }
    }\\
    \rule{\textwidth}{0.03cm}\\[0.2cm]
\end{center}
}

\definecolor{blue1}{rgb}{0.878431373,	0.921568627,	0.968627451}
\definecolor{blue2}{rgb}{0.654901961,	0.788235294,	0.870588235}
\definecolor{blue3}{rgb}{0.28627451,	0.505882353,	0.721568627}

\definecolor{LightCyan}{rgb}{0.88,1,1}
\definecolor{Blue}{rgb}{0, 0.5, 1}
\definecolor{Green}{rgb}{0.0, 0.8, 0.0 }
\definecolor{Red}{rgb}{0.95, 0.55, 0.6}
\definecolor{Skyblue}{rgb}{0.6, 0.6, 0.95 }
\definecolor{LightGray}{gray}{0.9}
\definecolor{Gray}{gray}{0.7}
\definecolor{ForestGreen}{rgb}{0.0, 0.4, 0.1}
\definecolor{Goldenrod}{rgb}{0.85, 0.64, 0.12}
\definecolor{Mulberry}{rgb}{0.46, 0.02, 0.215}
\definecolor{MidnightBlue}{rgb}{0.1, 0.1, 0.44}



\usepackage{framed}
\definecolor{shadecolor}{named}{LightGray}

\usepackage{times}
\usepackage{latexsym}

\usepackage[T1]{fontenc}

\usepackage[utf8]{inputenc}

\usepackage{microtype}

\usepackage{inconsolata}

\usepackage{graphicx}

\title{\mytitle}

\author{Siqi Liang \\ University of Michigan \\ Ann Arbor, MI, USA \\ \texttt{siqilian@umich.edu}
        \And
        Sumyeong Ahn \\ KENTECH \\ Naju, Jeollanam-do, Korea \\ \texttt{sumyeongahn@kentech.ac.kr}
        \AND
        Paramveer S. Dhillon \\ University of Michigan \\ Ann Arbor, MI, USA \\ \texttt{dhillonp@umich.edu}
        \And
        Jiayu Zhou\thanks{Corresponding author.} \\ University of Michigan \\ Ann Arbor, MI, USA \\ \texttt{jiayuz@umich.edu}}

\begin{document}
\maketitle
\begin{abstract}
In-context learning (ICL) relies heavily on high-quality demonstrations drawn from large annotated corpora. Existing approaches detect noisy annotations by ranking local perplexities, presuming that noisy samples yield higher perplexities than their clean counterparts. However, this assumption breaks down when the noise ratio is high and many demonstrations are flawed.
We re-examine the perplexity-based paradigm for text generation under noisy annotations, highlighting two sources of bias in perplexity: the annotation itself and the domain-specific knowledge inherent in large language models (LLMs). To overcome these biases, we introduce a dual-debiasing framework that uses synthesized neighbors to explicitly correct perplexity estimates, yielding a robust \emph{Sample Cleanliness Score}. This metric uncovers absolute sample cleanliness regardless of the overall corpus noise level.
Extensive experiments demonstrate our method’s superior noise-detection capabilities and show that its final ICL performance is comparable to that of a fully clean demonstration corpus. Moreover, our approach remains robust even when noise ratios are extremely high.
\end{abstract}
\section{Introduction}
Large Language Models (LLMs) have demonstrated impressive capabilities across a wide range of Natural Language Processing (NLP) tasks~\cite{brown2020language, touvron2023llama}. 
This performance is largely attributed to various techniques that leverage LLMs, such as Chain-of-Thought (CoT)~\cite{wei2022chain}, In-Context Learning (ICL)~\cite{dong-etal-2024-survey}, and so on. 
In particular, ICL guides LLMs by providing contextual examples to facilitate more accurate responses to queries. Typically, ICL involves two primary steps: (1) retrieving demonstration examples from a database that are relevant to the query, and (2) incorporating these samples as contextual input preceding the query. 
Numerous approaches have been proposed to enhance the retrieval process and consequently improve the quality of responses generated by LLMs~\citep{ye2023compositional,li2023unified}. 

\begin{figure*}[t]
    \centering
    \includegraphics[width=.95\textwidth]{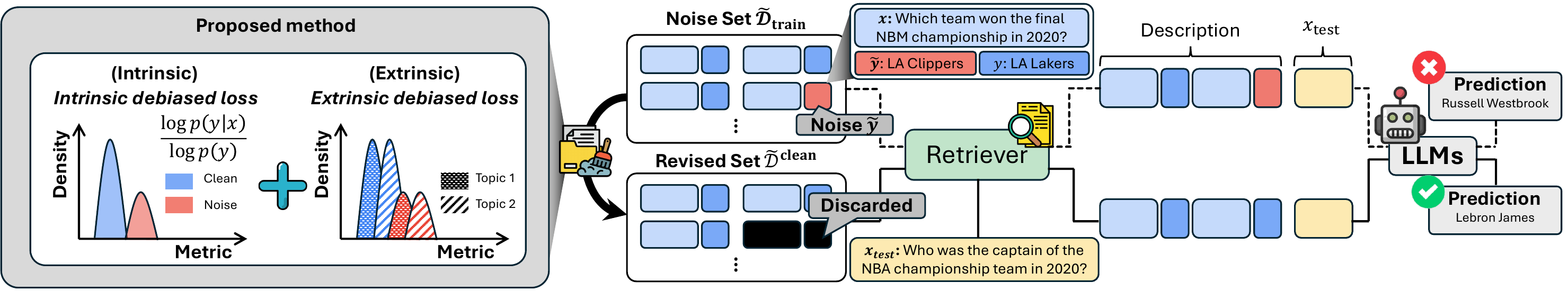}
    \caption{Overview of noisy ICL and the proposed method: When noisy information is present in an ICL dataset, it can lead to performance degradation (Top case). However, if the noisy information is sufficiently removed, performance can be improved (Bottom case). In this paper, we go beyond the conventional perplexity score-based approach and propose a Sample Cleanliness Score, which is based on intrinsic/extrinsic debiased loss.}
    \label{fig:intro}
\end{figure*}

However, nearly all research on ICL assumes that the underlying database of descriptions is entirely factual. 
Only a limited number of studies have addressed scenarios in which the database contains incorrect information, often referred to as noisy attributes. 
For instance, in~\citet{gao2024noise}, the authors proposed a local perplexity ranking method to identify noisy information and replace demonstrations deemed to be noisy. 
Similarly, the authors in~\citet{kang2024context} introduced an algorithm called {\it recitification}, which refines the labels by fine-tuning pre-trained LLMs, such as GPT-2~\cite{radford2019language}. 

While these approaches have explored this meaningful problem setting and shown robust results under demonstrations with noisy annotations, they still present certain limitations. First, the previous methods presuppose that clean demonstrations make up a substantial portion of the training set, leading to methodological failures when the noise ratio is high.
Second, the reliance on perplexity, or similar metrics, is predicated on the assumption that noisy demonstrations consistently manifest higher perplexity than clean ones. Lastly, existing studies have not attempted to articulate the influence of the LLM's prior knowledge on their ability to detect the matching relationship between query and annotation of demonstration samples.

In this paper, we begin by revisiting the naive probability-based metric, which depicts LLM's perception on the matching relationship between queries and annotations in the noisy ICL generation task.
We further investigate the potential influence of LLM's prior knowledge on the values of this metric, which we categorize as intrinsic bias and extrinsic bias. 
Based on our findings, we mathematically formulate both types of bias, providing a quantified assessment of the impact of LLM's prior knowledge on their ability to discern the matching relationship between query and annotation of each demonstration sample.

Building on these insights, we introduce an explicit dual-debiasing method that leverages neighbor-based feasible estimation to mitigate the effects of both intrinsic and extrinsic biases.
We then present the \textit{Sample Cleanliness Score}, a metric developed to detect noisy demonstrations, and our pipeline incorporating the score as a solution to the noisy challenge in ICL.

 We describe our key contributions as follows:
 \begin{itemize}[leftmargin=10pt]
     \item We reveal the potential influence of LLM's prior knowledge on their perception of the query-annotation matching relationship, specifically identifying \textbf{\textit{intrinsic bias}} and \textbf{\textit{extrinsic bias}}, and we provide their mathematical formulation.
     \item We propose practical approximations for intrinsic and extrinsic biases, utilizing a neighbor-based method for the latter, to enable effective assessments of these biases.
     \item We propose a novel dual-debiasing approach to mitigate the impact of LLM's prior knowledge on their perception of the query-annotation matching relationship. This method leads to developing the \textit{sample cleanliness score}, a new metric designed to detect noisy demonstrations.
     \item We design a metric-based pipeline for addressing the challenges of noisy ICL, which is sufficiently robust even under extreme noise cases.
     \item We evaluate the efficacy of our pipeline across diverse benchmark datasets for ICL text generation tasks under various noisy settings. Our results show superior performance compared with several baselines, with outcomes in many cases comparable to those achieved in clean settings.
 \end{itemize}
 \section{Preliminaries}
\label{sec:preliminaries}

We consider ICL in text generation tasks. 
Given the training set $\mathcal{D}_{\text{train}} = \{ (\boldsymbol{x}_i, \boldsymbol{y}_i) \}_{i=1}^N$, where $\boldsymbol{x}_i$ is the demonstration query text and $\boldsymbol{y}_i$ is the tokenized corresponding annotation with length $T_i=|\boldsymbol{y}_i|$, ICL aims to utilize a LLM to generate sequence output for test queries in the test set $\mathcal{D}_{\text{test}} = \{\boldsymbol{x}^{\text{test}}_j\}_{j=1}^M$. 
        
A typical ICL process contains retrieval and inference steps. 
Given a test query $\boldsymbol{x}^{\text{test}} \in \mathcal{D}_{\text{test}}$, the retriever retrieves $k$ demonstration examples from $\mathcal{D}_{\text{train}}$, say, $\mathcal{D}_{\text{ex}} = \{(\boldsymbol{x}_i, \boldsymbol{y}_i)\}_{i \in \mathcal{S}}$, where $\mathcal{S}$ is the index set of retrieved samples with $| \mathcal{S} |=k$. 
Then the prompt $\mathcal{P}$ will be constructed using the retrieved examples $\mathcal{D}_{\text{ex}}$ and the given test query $x^{\text{test}}$ based on the prompt template $\mathcal{T}$.
By feeding the constructed prompt $\mathcal{P}$ into LLM for inference, we can obtain generated results via:
\begin{align*}
    y_t \sim P_{\text{LLM}}(Y_t | \mathcal{P}, y_{<t}),
\end{align*}
where $\sim$ denotes the decoding strategies. 
ICL performance relies on the quality of retrieved demonstration examples~\citep{li2023unified, ye2023compositional}.

However, the training set in real-world settings can easily include noised annotations $\tilde{\mathcal{D}}_{\text{train}} = \{ (\boldsymbol{x}_i, \tilde{\boldsymbol{y}}_i) \}_{i=1}^N$, due to unreliable data sources or limited annotation expertise.
Thus, the retrieved demonstration examples $\tilde{\mathcal{D}}_{\text{ex}} = \{(\boldsymbol{x}_i, \tilde{\boldsymbol{y}}_i)\}_{i \in \mathcal{S}}$ can introduce misleading information in the prompt, which thus leads to degraded performance in ICL~\citep{yoo2022ground, gao2024noise}.

Though previous work by~\citet{gao2024noise} has identified the issue of noise in ICL for text generation tasks and suggested potential solutions, it presupposes that clean samples predominate in the training set. This leads to the failure of the proposed method when the noise ratio is high. 
Additionally, though the authors acknowledged the limitations of naive perplexity in distinguishing between noisy and clean demonstrations due to inherent perturbation, their method only mitigates this effect implicitly. 
It does not thoroughly examine the underlying mechanism of the query-annotation relationship for demonstration samples.
This gap in understanding motivates our further investigation into ICL generation tasks with noisy annotations to develop a more generalizable solution. Furthermore, it requires a deeper comprehension of the influence of LLM's parametric knowledge on its detection capability of noisy matching relations in query-annotation pairs.

\section{Method}

In this section, we will first introduce our metric design motivation. 
Then, we deliver the explicit dual-debiasing method to compute the \textit{Sample Cleanliness Score} for each demonstration sample from the given training set, including the intrinsic-debiasing step and the neighbor-based extrinsic-debiasing step. 
Finally, we introduce the complete pipeline for noisy ICL utilizing proposed \textit{Sample Cleanliness Score}. 

\subsection{Motivation on Probability}

Intuitively, a well-pretrained LLM is more likely to assign higher probability to the correct annotation (or, the in-distribution annotation) than the noised one (or, out-of-distribution one), when conditioned on the same query $\boldsymbol{x}$~\cite{arora2021types, alon2023detecting, gao2024noise},
that is, 
\begin{align*}
    P(\boldsymbol{y}^* | \boldsymbol{x}) > P(\tilde{\boldsymbol{y}} | \boldsymbol{x}),
\end{align*}
where $\boldsymbol{y}^*$ with length $T^*$ is the correct annotation for $\boldsymbol{x}$, and $\tilde{\boldsymbol{y}} \neq \boldsymbol{y}^*$ with length $\tilde{T}$ is the observed noised annotation.
Due to the impact of varied token sequence lengths, we consider the following token-wise version of the conditional probabilities: 
\begin{align}
\label{eq:root-cond-prob}
    P(\boldsymbol{y}^* | \boldsymbol{x})^{1/T^*} > P(\tilde{\boldsymbol{y}} | \boldsymbol{x})^{1/\tilde{T}}.
\end{align}
Now, applying the logarithmic transformation to both sides of~\autoref{eq:root-cond-prob}, we have:
\begin{align}
\label{eq:root-log-cond-prob}
    - \frac{1}{T^*} \log{P(\boldsymbol{y}^* | \boldsymbol{x})} < - \frac{1}{\tilde{T}}\log{P(\tilde{\boldsymbol{y}} | \boldsymbol{x})}.
\end{align}
Given $P(\boldsymbol{y} | \boldsymbol{x}) = \prod_{t=1}^T{P(y_t | \boldsymbol{x}, y_{<t})}$,
we define the following based on per-token conditioned probabilities for sequence $\boldsymbol{y}$ given prefix sequence $\boldsymbol{x}$:
\begin{align}
    \mathcal{L}( \boldsymbol{y} | \boldsymbol{x} ) = - \frac{1}{T} \sum\nolimits_{t=1}^{T}{\log{P(y_t | \boldsymbol{x}, y_{<t})}},
\end{align}
which is the per-token version of Negative Log-Likelihood (NLL) loss for $\boldsymbol{y}$ given prefix $\boldsymbol{x}$ and can be easily computed using next-token probabilities output from LLM.

\autoref{eq:root-log-cond-prob} can thus be rewritten as $ \mathcal{L}( \boldsymbol{y}^* | \boldsymbol{x} ) < \mathcal{L}( \tilde{\boldsymbol{y}} | \boldsymbol{x} )$ when $\tilde{\boldsymbol{y}} \neq \boldsymbol{y}^*$.
This suggests that for a fixed query token sequence $\boldsymbol{x}$, the noised annotation is expected to exhibit a higher per-token NLL loss value than its clean counterpart.

However, directly using $\mathcal{L}(\boldsymbol{y} | \boldsymbol{x})$ does a poor job on differentiating noisy and clean demonstrations. 
We provide this analysis in~\autoref{app:analysis_perplexity}. 
In short, the distribution of clean demonstrations' per-token NLL loss values overlaps heavily with that of noisy ones, which makes it almost impossible to determine whether a demonstration is clean or not given its $\mathcal{L}(\boldsymbol{y} | \boldsymbol{x})$ value alone. 

This can be attributed to bias factors like the demonstration sample itself, LLM's prior knowledge~\citep{fei2023mitigating,zhao2021calibrate}, and even LLM architectures~\citep{o2023contrastive,li2022contrastive}. 
The previous paper~\citep{gao2024noise} also noticed a similar phenomenon on sample-wise perplexity. 
Unlike their method, which tries to disentangle the perplexity impact of LLMs implicitly, we propose the dual-debiasing method to remove the bias derived from the demonstration itself, which we call \textbf{\textit{intrinsic debiasing}}, as well as the bias derived from various expertise levels of LLM's parametric knowledge on different domains, which we call \textbf{\textit{extrinsic debiasing}}. 
Our ultimate objective is to formulate a metric for each demonstration sample $(\boldsymbol{x}_i, \tilde{\boldsymbol{y}}_i) \sim \tilde{\mathcal{D}}_\text{train}$  that satisfies two key properties:\quad 1) it accesses the \textit{matching relationship between the annotation and the query} of each demonstration;\quad 2) it is \textit{comparable across different demonstration samples}, regardless of variations in both query $\boldsymbol{x}_i$ and annotation $\tilde{\boldsymbol{y}}_i$. 
We intend for this metric to facilitate the determination of whether a demonstration is clean or noisy based on its metric value.

\subsection{Intrinsic Debiasing}



When we only consider $\mathcal{L}(\tilde{\boldsymbol{y}} | \boldsymbol{x})$ to evaluate the matching relationship between $\boldsymbol{x}$ and observed $\tilde{\boldsymbol{y}}$, the pre-trained LLM may be very familiar with $\tilde{\boldsymbol{y}}$ itself, given the frequent occurrence of $\tilde{\boldsymbol{y}}$ in the pre-training dataset. 
This will lead to $\mathcal{L}(\tilde{\boldsymbol{y}} | \boldsymbol{x}) < \mathcal{L}(\boldsymbol{y}^* | \boldsymbol{x})$ even when the observed $\tilde{\boldsymbol{y}}$ does not match with $\boldsymbol{x}$ as noised annotation, given LLM assigns high probability on $\tilde{\boldsymbol{y}}$ than $\boldsymbol{y}^*$ without any prefix token sequence. 
In other words, the naive $\mathcal{L}(\boldsymbol{y} | \boldsymbol{x})$ is biased by LLM's prior knowledge on annotation $\boldsymbol{y}$. 
Since this bias is derived from the annotation part of the demonstration sample itself, we name it \textit{intrinsic bias}.

Motivated by this, we propose the \textit{\textbf{intrinsic debiasing}} step to remove LLM's prior knowledge bias on $\boldsymbol{y}$, i.e., $P(\boldsymbol{y})$. 
We define the per-token loss function for sequence $\boldsymbol{y}$ \emph{without any prefix} as:
\begin{align*} 
    \mathcal{L}(\boldsymbol{y}) = - \frac{1}{T} \sum\nolimits_{t=1}^{T}{\log{P(y_t | y_{<t})}},
\end{align*}
which serves as an effective alternative for representing $P(\boldsymbol{y})$.
Then we defined the intrinsic-debiased per-token loss function as:
\begin{align}
\label{eq:int-debias-definition}
\mathcal{L}_{\text{de-int}}(\boldsymbol{y}|\boldsymbol{x}) = \frac{\mathcal{L}( \boldsymbol{y} | \boldsymbol{x} )}{\mathcal{L}( \boldsymbol{y})}.
\end{align}
Given a fixed demonstration query $\boldsymbol{x}$, a clean annotation is expected to exhibit a lower value of $\mathcal{L}_{\text{de-int}}(\boldsymbol{y}|\boldsymbol{x})$ than a noised one. 
Specifically, if $\tilde{\boldsymbol{y}} \neq \boldsymbol{y}^*$, then
\begin{align}
\label{eq:int-debias-comparison}
    \mathcal{L}_{\text{de-int}}(\tilde{\boldsymbol{y}} |\boldsymbol{x}) > \mathcal{L}_{\text{de-int}}(\boldsymbol{y}^* | \boldsymbol{x}).
\end{align}
Furthermore, in cases where the ground-truth annotation is unavailable and only two observed annotations, $\tilde{\boldsymbol{y}}_1$ and $\tilde{\boldsymbol{y}}_2$, are provided for a given $\boldsymbol{x}$, it can be inferred that $\tilde{\boldsymbol{y}}_1$ is more mismatching with $\boldsymbol{x}$ than $\tilde{\boldsymbol{y}}_2$ if $\mathcal{L}_{\text{de-int}}(\tilde{\boldsymbol{y}}_1 |\boldsymbol{x}) > \mathcal{L}_{\text{de-int}}(\tilde{\boldsymbol{y}}_2 | \boldsymbol{x})$.



\subsection{Neighbor-based Extrinsic Debiasing}
\label{sec:neighbor_extrinsic_debias}

Using $\mathcal{L}_{\text{de-int}}$, we can assess the relative query-annotation mismatch between two demonstrations that share the same query $\boldsymbol{x}$.
However, within the dataset $\tilde{\mathcal{D}}_{\text{train}}$, each query is associated with only one annotation - either clean or noisy, but never both simultaneously.
This raises an important question: 
Are the intrinsic-debiased per-token loss values comparable when both queries and annotations differ? Specifically, given pairs $(\boldsymbol{x}_i, \tilde{\boldsymbol{y}}_i)$ and $(\boldsymbol{x}_j, \tilde{\boldsymbol{y}}_j)$ with $\boldsymbol{x}_i \neq \boldsymbol{x}_j$ and $\tilde{\boldsymbol{y}}_i \neq \tilde{\boldsymbol{y}}_j$, can we conclude that $(\boldsymbol{x}_i, \tilde{\boldsymbol{y}}_i)$ is more likely to be noisy than $(\boldsymbol{x}_j, \tilde{\boldsymbol{y}}_j)$ if $\mathcal{L}_{\text{de-int}}(\tilde{\boldsymbol{y}}_i | \boldsymbol{x}_i) > \mathcal{L}_{\text{de-int}}(\tilde{\boldsymbol{y}}_j | \boldsymbol{x}_j)$?

\begin{figure}[t]
    \centering
    \subfloat[\textit{Location} vs \textit{Person} on \\ \centering{\code{Gemma-2b}}\label{fig:int-de-good1}]{
        \includegraphics[width=0.42\columnwidth]{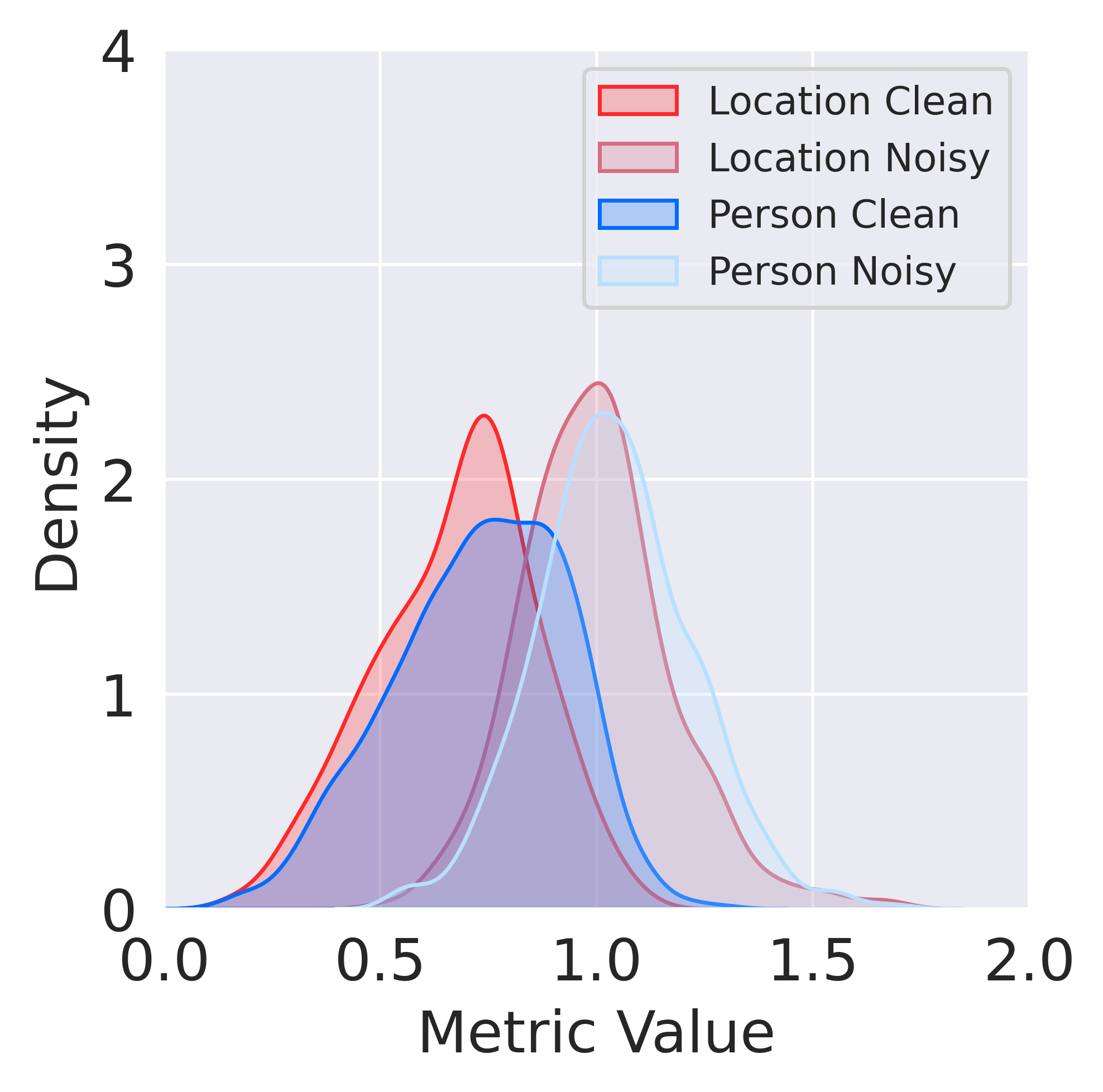}
    }
    \hspace{2mm}
    \subfloat[\textit{Date} vs \textit{Person} on \\ \centering{\code{GPT-Neo-2.7B}}\label{fig:int-de-good2}]{
        \includegraphics[width=0.42\columnwidth]{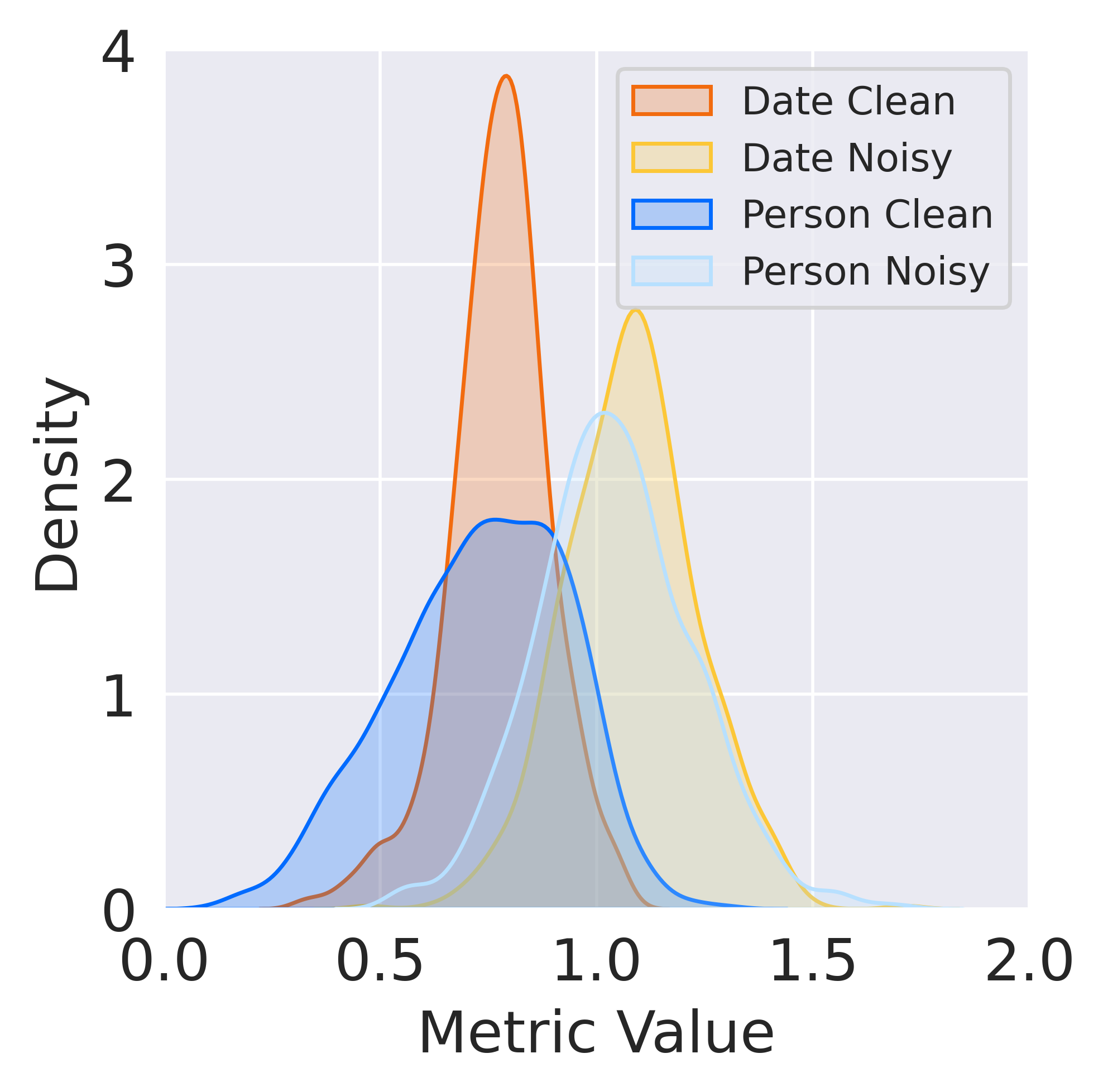}
    }
    \hfill
    \subfloat[\textit{Date} vs \textit{Sports team} on \\ \centering{\code{Gemma-2b}}\label{fig:int-de-bad1}]{
        \includegraphics[width=0.42\columnwidth]{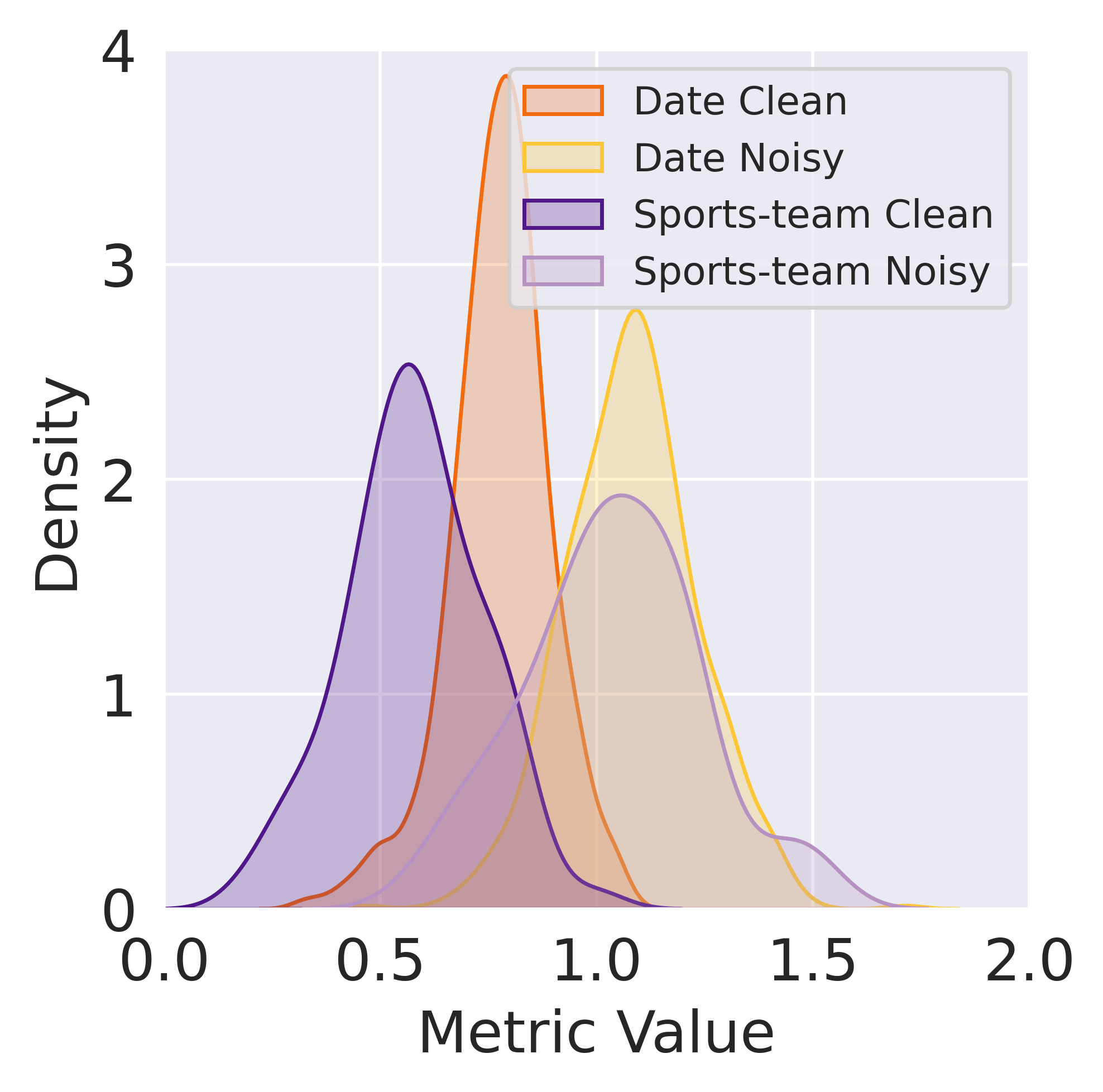}
    }
    \hspace{2mm}
    \subfloat[\textit{Band} vs \textit{Number} on \\ \centering{\code{GPT-Neo-2.7B}}\label{fig:int-de-bad2}]{
        \includegraphics[width=0.42\columnwidth]{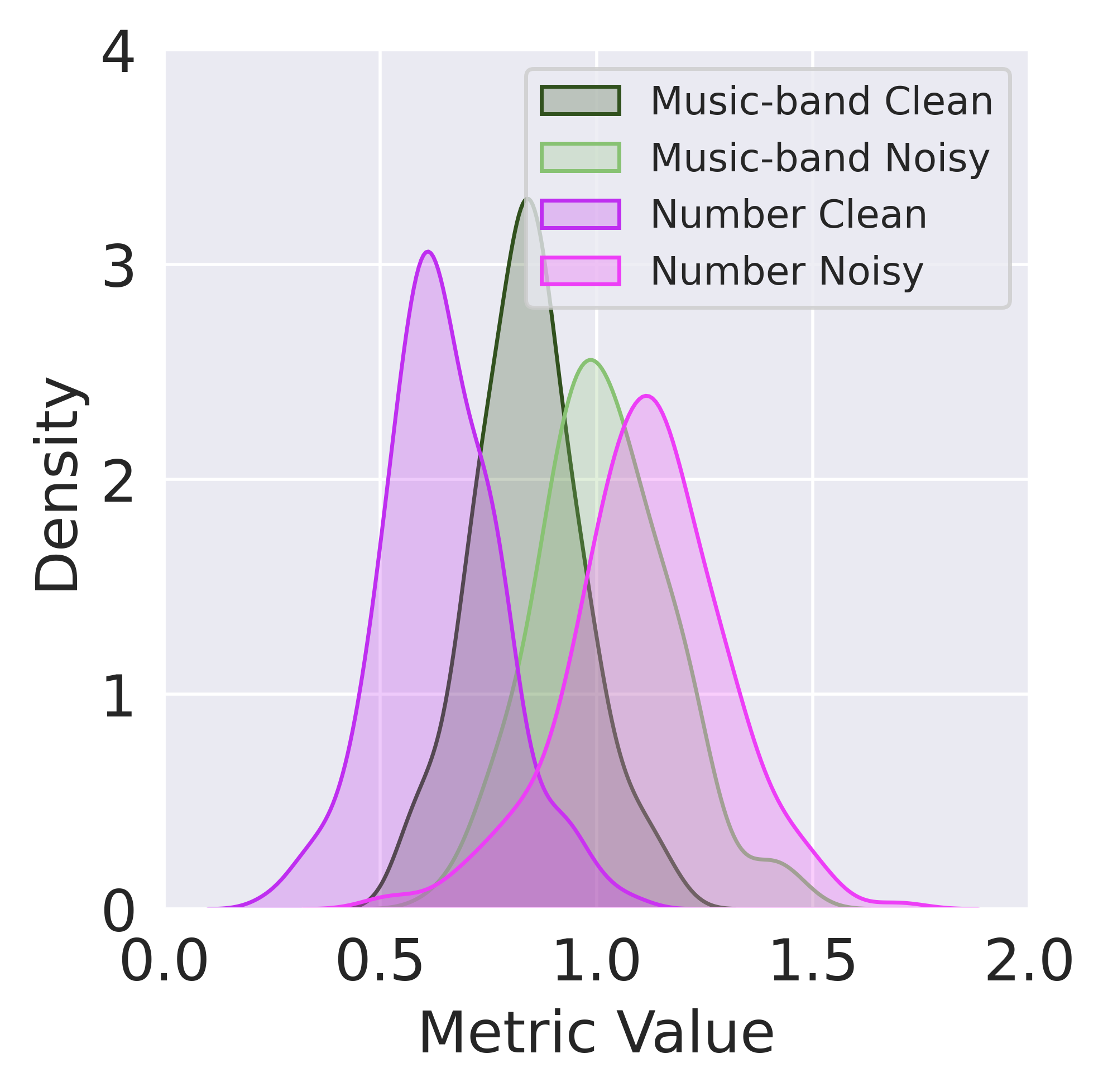}
    }
    \caption{Distribution of $\mathcal{L}_{\text{de-int}}$ for both clean and noisy versions of demonstrations for different topics. The $\mathcal{L}_{\text{de-int}}$ values are calculated using \code{Gemma-2b} and \code{GPT-Neo-2.7B}.}
    \vspace{-5pt}
    \label{fig:int-de-topic-dist}
\end{figure} LLMs exhibit varying levels of expertise across different knowledge domains. 
For instance, GPT-4~\citep{achiam2023gpt} exhibits diverse capabilities in translation tasks, as demonstrated in recent benchmark studies~\citep{yan2024benchmarking}. 
Moreover, LLMs show different degrees of familiarity with queries from distinct knowledge domains, which affects the value distributions of $\mathcal{L}_{\text{de-int}}$ for samples from these domains. 
Consequently, the $\mathcal{L}_{\text{de-int}}$ values of different domains are not directly comparable, making it challenging to compare samples such as $(\boldsymbol{x}_i, \tilde{\boldsymbol{y}}_i)$ and $(\boldsymbol{x}_j, \tilde{\boldsymbol{y}}_j)$ when they originate from different knowledge domains.

Motivated by this, we categorize the training set of NQ~\citep{kwiatkowski2019natural} dataset into different sample topics using GPT-4, and visualize the $\mathcal{L}_{\text{de-int}}$ value distributions for both noisy and clean demonstration samples across different topics, including \textit{Location}, \textit{Person}, \textit{Date}, \textit{Music-Band}, \textit{Number} and \textit{Sports-Team}. We use both \code{Gemma-2b}~\citep{team2024gemma} and \code{GPT-Neo-2.7B}~\citep{gpt-neo} to calculate the $\mathcal{L}_{\text{de-int}}$ values for demonstrations. 

As shown in~\autoref{fig:int-de-good1} and \autoref{fig:int-de-good2}, within each topic, the value distribution of noisy demonstrations is significantly separated from that of clean ones, and clean demonstrations maintain similar value distributions across different topics. Specifically, \autoref{fig:int-de-good1} reveals that clean/noisy demonstrations from \textit{Location} share similar distributions with clean/noisy demonstrations from \textit{Person} under \code{Gemma-2b}. Similarly, \autoref{fig:int-de-good2} shows this pattern for \textit{Date} vs \textit{Person} under \code{GPT-Neo-2.7B}. 
In these cases, we can directly compare $\mathcal{L}_{\text{de-int}}$ values between demonstrations from different topics to determine their relative query-annotation mismatching levels.

However, different patterns emerge when examining demonstrations from \textit{Date} vs \textit{Sports-Team} under \code{Gemma-2b}, and \textit{Music-Band} vs \textit{Number} under \code{GPT-Neo-2.7B}. 
As shown in~\autoref{fig:int-de-bad1}, under \code{Gemma-2b}, while clean and noisy demonstrations are well-separated within each topic, the mean value of \textit{Date} clean demonstration distribution is significantly higher than that of \textit{Sports-Team} clean demonstrations, leading to more distribution overlap with \textit{Sports-Team} noisy demonstrations. 
This distribution pattern can lead to false noise detection for \textit{Date} clean demonstrations, as they may be incorrectly classified as noisy when their $\mathcal{L}_{\text{de-int}}$ values are compared with \textit{Sports-Team} ones. We observe a similar issue between \textit{Music-Band} vs \textit{Number} under \code{Gemma-2b} as shown in~\autoref{fig:int-de-bad2}. 

To conclude, LLMs' bias across different knowledge domains makes $\mathcal{L}_{\text{de-int}}$ values incomparable for demonstration samples from different domains. 
We term this domain-related bias as \textit{extrinsic bias} since it originates from the demonstration sample's domain rather than the sample itself.

Based on these observations, we propose the \textbf{\textit{extrinsic debiasing}} step to eliminate bias associated with different knowledge domains. 
This step aims to facilitate comparability of the metric values across demonstrations, even when they originate from distinct domains. 
Given data point~\footnote{The definition of $(\boldsymbol{x}, \boldsymbol{y})$ here is not on the discrete demonstration sample space, but in the continuous space.} $(\boldsymbol{x}, \boldsymbol{y}) \in \mathcal{X} \times \mathcal{Y}$, we define an associated domain based on a metric space $(\mathcal{X} \times \mathcal{Y}, d)$ with a distance function $d$:
\begin{align}
\mathcal{N}\left((\boldsymbol{x}, \boldsymbol{y})\right) = \big\{ 
& (\boldsymbol{x}^{\prime}, \boldsymbol{y}^{\prime}) \in \mathcal{X} \times \mathcal{Y}: \notag \\
& d\left((\boldsymbol{x}, \boldsymbol{y}), (\boldsymbol{x}^\prime, \boldsymbol{y}^\prime)\right) < \eta \big\},
\end{align}
where $d$ is a suitable distance metric, and $\eta$ is a positive real number indicating the radius of the associated domain. 
For ease of notation and where the context is clear, we will subsequently denote $\mathcal{N}\left((\boldsymbol{x}, \boldsymbol{y})\right)$ simply as $\mathcal{N}$. 
Assume the density function $\rho$ is uniform over given defined domain $\mathcal{N}\left((\boldsymbol{x}, \boldsymbol{y})\right)$ with volume $V$:
\begin{align*}
\rho\left(\boldsymbol{x}^{\prime}, \boldsymbol{y}^{\prime}\right)= \frac{1}{V}, \; \text { if } \; (\boldsymbol{x}^{\prime}, \boldsymbol{y}^{\prime}) \in \mathcal{N}\left((\boldsymbol{x}, \boldsymbol{y})\right),
\end{align*}
where $V$ is the measure of the associate domain defined by $V=\int_{(\boldsymbol{x}^{\prime}, \boldsymbol{y}^{\prime}) \in \mathcal{N}} d V$ and $d V$ is the differential volume element in the continuous space $\mathcal{X} \times \mathcal{Y}$.
To this end, we can estimate the \textit{extrinsic bias}, denoted as $\Phi\left((\boldsymbol{x}, \boldsymbol{y})\right)$, over the domain of $(\boldsymbol{x}, \boldsymbol{y})$ using the empirical estimation of $\mathcal{L}_{\text{de-int}}$: 
\begin{align}
\label{eq:continuous-extrinsic-bias}
    \Phi\left((\boldsymbol{x}, \boldsymbol{y})\right) = \frac{1}{V} \int_{(\boldsymbol{x}^{\prime}, \boldsymbol{y}^{\prime}) \in \mathcal{N}} {\mathcal{L}_{\text{de-int}}(\boldsymbol{y}^{\prime} | \boldsymbol{x}^{\prime})} d V.
\end{align}
Then we can perform the extrinsic debiasing via $\mathcal{L}_{\text{de-int}}(\boldsymbol{y}|\boldsymbol{x}) / \Phi\left((\boldsymbol{x}, \boldsymbol{y})\right)$,
which equivalently, we introduce the final metric used for each demonstration sample, \ie \textit{Sample Cleanliness Score}, as  
\begin{align}
\label{eq:scs_def}
    \mathcal{I}(\boldsymbol{x},\boldsymbol{y}) = \frac{\Phi\left((\boldsymbol{x}, \boldsymbol{y})\right)}{\mathcal{L}_{\text{de-int}}(\boldsymbol{y}|\boldsymbol{x})}.
\end{align}

Given two demonstration samples $(\boldsymbol{x}_1, \tilde{\boldsymbol{y}}_1)$ and $(\boldsymbol{x}_2, \tilde{\boldsymbol{y}}_2)$ from $\tilde{\mathcal{D}}_{\text{train}}$, if $\tilde{\boldsymbol{y}}_1$ is clean while $\tilde{\boldsymbol{y}}_2$ is noised, the following relation holds
\begin{align*}
    \mathcal{I}(\boldsymbol{x}_1,\tilde{\boldsymbol{y}}_1) > \mathcal{I}(\boldsymbol{x}_2, \tilde{\boldsymbol{y}}_2),
\end{align*}
even when two samples are from different domains.

\begin{figure}[H]
        \centering
        \includegraphics[width=1.0\columnwidth]{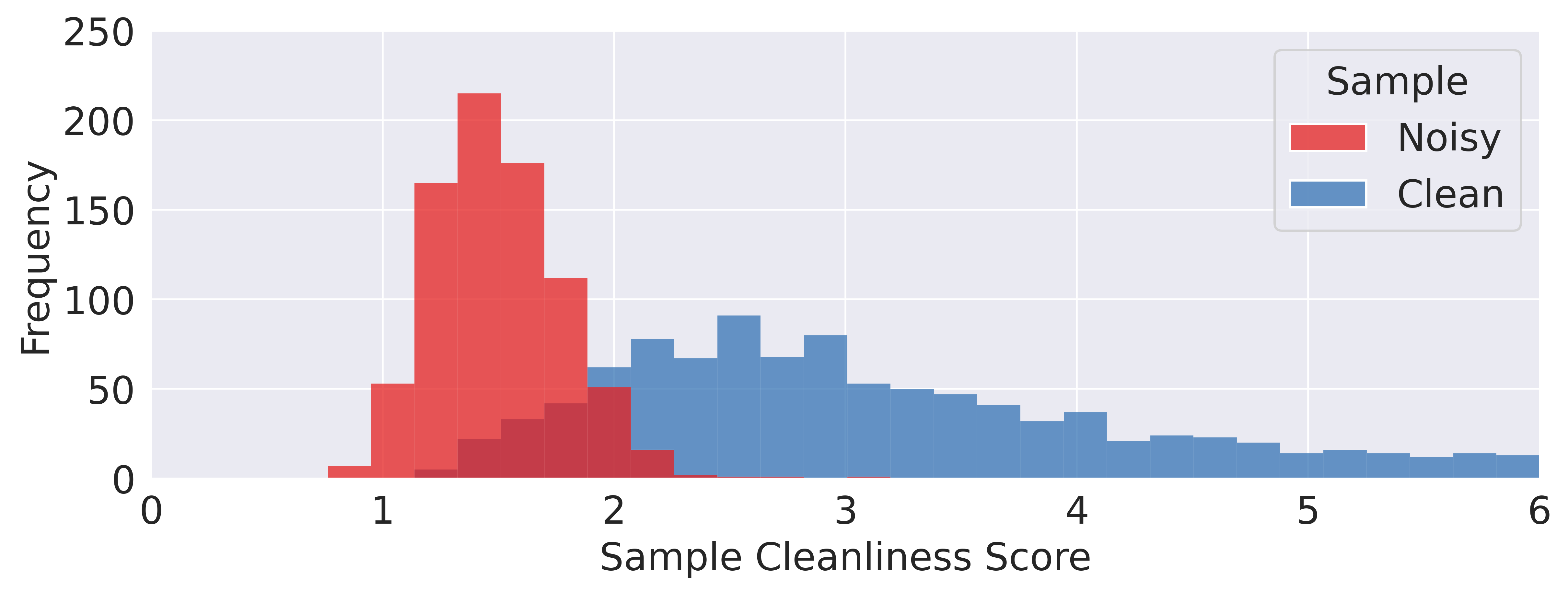}    
        \caption{Sample Cleanliness Score distributions on noisy NQ training set. A larger value indicates that it is more likely to be a clean sample. It shows the pattern of distinguishability between noisy and clean samples.}
        \label{fig:domain-debias}
\end{figure}

Note that \autoref{eq:continuous-extrinsic-bias} is defined on the continuous space, which is intractable to estimate. Thus, we propose to use the finite discrete neighboring demonstration samples to provide a tractable estimation. We provide the following construction of neighboring demonstration samples to serve as alternation for the domain $\mathcal{N}$:
\begin{align}
\mathcal{N}_{\text{DISC}}\left((\boldsymbol{x}, \boldsymbol{y})\right) = \big\{ &(\boldsymbol{x}, \boldsymbol{y}_z^{\prime}) \big\}_{z=1}^{N_{\text{neighbor}}},
\end{align}
where $\boldsymbol{y}_z^\prime$ can be tokenized sequence sampled from a large corpus $\mathcal{C}$, and $N_{\text{neighbor}}$ is the number of neighbors. We denote $\mathcal{N}_{\text{DISC}}\left((\boldsymbol{x}, \boldsymbol{y})\right)$ simply as $\mathcal{N}_{\text{DISC}}$ for ease of notation.
We can define the distance function $d(\cdot, \cdot)$ based on Edit Distance $d_{\text{edit}}$, which can help us to bound the radius of $\mathcal{N}_{\text{DISC}}$: 
\begin{align*}
    d((\boldsymbol{x}, \boldsymbol{y}), (\boldsymbol{x}^\prime, \boldsymbol{y}^\prime)) 
    &= d((\boldsymbol{x}, \boldsymbol{y}), (\boldsymbol{x}, \boldsymbol{y}^\prime)) \\
    &= d_{\text{edit}}( \boldsymbol{y},  \boldsymbol{y}^\prime) \\
    & \leq \max{(T, T_{\text{max}})} = \eta \\
    & \text{ for } \forall (\boldsymbol{x}^\prime, \boldsymbol{y}^\prime) \in \mathcal{N}_{\text{DISC}}\left((\boldsymbol{x}, \boldsymbol{y})\right),
\end{align*}
where $T_{\text{max}}$ is the maximum length of sequences in $\mathcal{C}$. 
And \autoref{eq:continuous-extrinsic-bias} can be replaced by:
\begin{align}
\label{eq:extrinsic_bias_estimate}
\Phi\left((\boldsymbol{x}, \boldsymbol{y})\right) = \frac{\sum_{(\boldsymbol{x}^\prime, \boldsymbol{y}^\prime) \sim \mathcal{N}_{\text{DISC}}}{\mathcal{L}_{\text{de-int}}(\boldsymbol{y}^\prime|\boldsymbol{x}^\prime)}}{N_{\text{neighbor}}}.
\end{align}
Consequently, \textit{Sample Cleanliness Score} $\mathcal{I}(\boldsymbol{x}_i, \tilde{\boldsymbol{y}}_i)$ is easy to calculate using neighbor-based extrinsic debiasing step for each $(\boldsymbol{x}_i, \tilde{\boldsymbol{y}}_i) \in \tilde{D}_{\text{train}}$. 
We simply use $\mathcal{I}_i$ for $\mathcal{I}(\boldsymbol{x}_i, \tilde{\boldsymbol{y}}_i)$ for short if context is clear.
We depict the distribution of the proposed $\mathcal{I}$ for both clean and noisy samples in~\autoref{fig:domain-debias}. 
The results show that we can now effectively differentiate noisy samples from clean ones based on the metric values.

\subsection{Complete Noisy ICL Pipeline}
\label{sec:noisy_icl_pipeline}

We now introduce our complete noisy ICL pipeline based on the proposed \textit{Sample Cleanliness Score}. 
The pipeline contains three steps: 
1) noisy demonstration detection based on the proposed metric; 
2) noisy demonstration cleanse; 
3) regular ICL process.
We provide detailed pseudocode for our noisy ICL pipeline in~\autoref{alg:pseudocode} of Appendix C.

Our initial step involves the Gaussian Mixture Model (GMM) to detect noisy demonstrations following the classic noisy label learning routine~\citep{lidividemix,huang2023twin,karim2022unicon}. 
We fit a 2-component GMM on $\left\{\mathcal{I}_i\right\}_{i=1}^N$, calculated for samples of the whole demonstration set, using the Expectation-Maximization algorithm. 
Then calculate the noisy probability of each demonstration sample $q_i$ as the posterior probability $q(g_{\text{noisy}} \mid \mathcal{I}_i)$, that indicates the possibility of $\mathcal{I}_i$ belongs to the Gaussian component $g_{\text{noisy}}$ representing the noisy demonstration samples, with a lower mean value of \textit{Sample Cleanliness Score}. 
Using threshold $\gamma$ on $q_i$ on all demonstration samples to separate the whole training set $\tilde{\mathcal{D}}_{\text{train}}$ into clean subset $\tilde{\mathcal{D}}^{\text{clean}}$ and noisy subset $\tilde{\mathcal{D}}^{\text{noisy}}$.

The second step is to handle the detected clean subset $\tilde{\mathcal{D}}^{\text{clean}}$ and noisy subset $\tilde{\mathcal{D}}^{\text{noisy}}$ obtained from previous step. 
One can follow \cite{gao2024noise} to replace each detected noisy demonstration with the nearest clean ones from $\tilde{\mathcal{D}}^{\text{clean}}$.
Yet the powerful denoising capability of the proposed metric suggests an alternative approach that eliminates all identified noisy samples, which achieves efficiency by reducing the size of the final training set for sample retrieval. 
Our experimental results show that the simple strategy achieves surprising ICL performance, even comparable to a clean setting.

The concluding phase of our pipeline entails the standard ICL process, which encompasses retrieval and inference, as outlined in~\autoref{sec:preliminaries}. 
However, in this phase, the original training set for retrieval is substituted with the identified clean subset $\tilde{\mathcal{D}}^{\text{clean}}$.

\section{Experiment}

\subsection{Experiment Setting}

\paragraph{Datasets:} We evaluate the performance of the proposed algorithm using four different datasets: 
(1) Natural Questions (\textbf{NQ}): a large-scale dataset containing real user queries/ from Google Search, paired with human-annotated answers. 
(2) Web Questions (\textbf{WebQ}): a dataset consisting of questions posed to Google Search, with answers derived from Freebase. 
(3) \textbf{SciQ}: a science-focused dataset with multiple-choice questions covering physics, chemistry, and biology. 
(4) \textbf{SQuAD}: a reading comprehension dataset containing questions and answers based on Wikipedia passages. 

\paragraph{Language Models:} We use \code{Llama-2-7B}~\citep{touvron2023llama} as the default ICL inference LLM. 
For the calculation of metrics, we use \code{Llama-2-7B} as the default model, and use \code{GPT-Neo-1.3B}~\citep{gpt-neo}, 
\code{Gemma-2b}~\citep{team2024gemma} and \code{Mistral-7B-v0.1}~\citep{jiang2023mistral} for analysis experiments.

\paragraph{Implementation details:} 
We implement our noisy ICL pipeline and the baselines based on OpenICL~\citep{wu2023openicl}. 
We use Random, TopK, and DPP~\citep{ye2023compositional} retrievers for the retrieval process. We follow \citet{gao2024noise} for the noise generation process for both relevant/irrelevant noise.
And the default hyper-parameters for noise detection setting are $\gamma=0.5$, $N_{\text{neighbor}}=50$. 
In the construction of $\mathcal{N}_{\text{DISC}}$, we deploy two implementations by utilizing two distinct corpora $\mathcal{C}$, for the sampling of $\boldsymbol{y}^\prime$. 
The first is termed the in-distribution corpus $\mathcal{C}_{\text{in}}$ which is compiled from all annotations within the observed $\tilde{\mathcal{D}}_{\text{train}}$; 
the second is designated as the out-distribution corpus, $\mathcal{C}_{\text{out}}$, and is comprised of annotations from external datasets.
To enhance the estimation of extrinsic bias, we constraint $T_{\text{max}}$ to equal the maximum annotation length in $\tilde{\mathcal{D}}_{\text{train}}$, thereby bounding the radius $\eta$ of demonstration sample's associated domain to $T_{\text{max}}$. We provide a detailed description of the generation process of neighbors in~\autoref{app:neighbor_generation}.
For the main results, we report the optimal performance achieved by either $\mathcal{C}_{\text{in}}$ or $\mathcal{C}_{\text{out}}$. 
However, $\mathcal{C}_{\text{out}}$ is exclusively used as the default implementation for analytical experiments. 
Furthermore, to save the computation costs, we employ a fixed set of $\boldsymbol{y}^\prime$ for all demonstration samples within the same dataset, thereby decreasing the computational expenditure required for estimating the extrinsic bias $\Phi$.

\vspace{-0.1cm}
\paragraph{Baselines:}
We evaluate the performance of our framework alongside three baselines: 
(1) Naive ICL: the conventional ICL pipeline without employing any specialized cleansing method, meaning that noisy information may be included in the description. 
(2) Random delete: a method that removes a randomly selected subset of samples corresponding to the noise ratio. 
(3) LPR~\cite{gao2024noise}: a method that leverages perplexity to cleanse the description using a local perplexity ranking score. 
(4) Ours: the proposed cleansing method utilizing the dual-debiasing approach.

\begin{table*}[h!]
\centering
\resizebox{1.0\textwidth}{!}{
\begin{tabular}{lllccccccccclllccccccccc}
\thickhline
\multirow{2}{*}{\textbf{Dataset}} & \multirow{2}{*}{\textbf{Retriever}} & \multirow{2}{*}{\textbf{Method}} 
& \multirow{2}{*}{\textbf{Clean}}
& \multicolumn{4}{c}{\textbf{Irrelevant Noise}} 
& \multicolumn{4}{c}{\textbf{Relevant Noise}} 
& \multirow{2}{*}{\textbf{Dataset}} & \multirow{2}{*}{\textbf{Retriever}} & \multirow{2}{*}{\textbf{Method}} 
& \multirow{2}{*}{\textbf{Clean}}
& \multicolumn{4}{c}{\textbf{Irrelevant Noise}} 
& \multicolumn{4}{c}{\textbf{Relevant Noise}} \\
\cmidrule(lr){5-8} \cmidrule(lr){9-12}
\cmidrule(lr){17-20} \cmidrule(lr){21-24}
& & & 
& \textbf{0.2} & \textbf{0.4} & \textbf{0.6} & \textbf{0.8} 
& \textbf{0.2} & \textbf{0.4} & \textbf{0.6} & \textbf{0.8} 
& & & &
& \textbf{0.2} & \textbf{0.4} & \textbf{0.6} & \textbf{0.8} 
& \textbf{0.2} & \textbf{0.4} & \textbf{0.6} & \textbf{0.8} \\
\midrule
\multirow{12}{*}{NQ}
& \multirow{4}{*}{Random} 
& Na\"ive ICL         & \textbf{14.00}
                    & 12.07         & \phantom{0}9.20   & \phantom{0}6.87   & \phantom{0}3.67
                    & 12.13         & 11.87             & 10.60             & \phantom{0}5.33 
& \multirow{12}{*}{SciQ} 
& \multirow{4}{*}{Random} 
& Na\"ive ICL         & 74.83
                    & 68.74         & 56.84             & 39.83             & 20.29
                    & 73.62         & 71.44             & 65.29             & 56.38 \\
&
& Random            & -
                    & 11.73         & \phantom{0}9.53   & \phantom{0}6.53   & \phantom{0}4.53 
                    & 13.13         & 12.40             & 10.40             & \phantom{0}8.67 
&&  
& Random            & - 
                    & 68.79         & 58.79             & 40.57             & 22.64 
                    & 73.10         & 70.75             & 66.15             & 57.13 \\
&
& LPR               & 12.80
                    & 11.67         & 10.47             & \phantom{0}8.47   & \phantom{0}5.33 
                    & 11.53         & 10.33             & \phantom{0}9.73   & \phantom{0}7.33 
&&
& LPR               & 69.43 
                    & 64.60         & 54.77             & 40.06             & 20.92 
                    & 66.38         & 61.90             & 56.95             & 44.60 \\
&
& \cellcolor{LightCyan} Ours              & \cellcolor{LightCyan} 13.40
                    & \cellcolor{LightCyan} \textbf{13.47}        & \cellcolor{LightCyan} \textbf{13.93}             & \cellcolor{LightCyan} \textbf{14.00}             & \cellcolor{LightCyan} \textbf{13.60}
                    & \cellcolor{LightCyan} \textbf{13.93}        & \cellcolor{LightCyan} \textbf{13.33}             & \cellcolor{LightCyan} \textbf{13.73}             & \cellcolor{LightCyan} \textbf{12.33} 
&&
& \cellcolor{LightCyan} Ours              & \cellcolor{LightCyan} \textbf{75.98}
                    & \cellcolor{LightCyan} \textbf{75.23}         & \cellcolor{LightCyan} \textbf{75.98}             & \cellcolor{LightCyan} \textbf{74.77}             & \cellcolor{LightCyan} \textbf{74.83} 
                    & \cellcolor{LightCyan} \textbf{75.46}         & \cellcolor{LightCyan} \textbf{75.17}             & \cellcolor{LightCyan} \textbf{76.03}             & \cellcolor{LightCyan} \textbf{75.63} \\
\cmidrule(lr){2-12} \cmidrule(lr){14-24}
& \multirow{4}{*}{TopK} 
& Na\"ive ICL         & \textbf{17.40}
                    & 15.07         & 10.80             & \phantom{0}9.33   & \phantom{0}5.33 
                    & 15.93         & 13.53             & 12.33             & \phantom{0}7.33 
&
& \multirow{4}{*}{TopK} 
& Na\"ive ICL         & 74.14 
                    & 67.18         & 51.78             & 36.95             & 18.56 
                    & 71.44         & 64.20             & 58.45             & 48.22 \\
&
& Random     & -
                    & 15.47         & 11.20             & \phantom{0}8.67   & \phantom{0}5.27
                    & 16.20         & 13.67             & 11.53             & \phantom{0}8.27 
&&  
& Random     & - 
                    & 68.05         & 54.02             & 38.16             & 20.75 
                    & 72.01         & 67.07             & 60.46             & 52.24 \\
&
& LPR               & 12.73
                    & 12.84         & 11.87             & 10.33             & \phantom{0}6.67
                    & 12.67         & 11.13             & \phantom{0}9.47   & \phantom{0}7.40 
&&
& LPR               & 69.20 
                    & 64.14         & 57.01             & 41.21             & 22.82 
                    & 66.38         & 62.64             & 53.28             & 40.11 \\
&
& \cellcolor{LightCyan} Ours     & \cellcolor{LightCyan} 16.33 
                    & \cellcolor{LightCyan} \textbf{16.40}        & \cellcolor{LightCyan} \textbf{16.33}             & \cellcolor{LightCyan} \textbf{15.93}             & \cellcolor{LightCyan} \textbf{15.40} 
                    & \cellcolor{LightCyan} \textbf{16.00}        & \cellcolor{LightCyan} \textbf{15.60}             & \cellcolor{LightCyan} \textbf{15.87}             & \cellcolor{LightCyan} \textbf{13.80} 
&&
& \cellcolor{LightCyan} Ours     & \cellcolor{LightCyan} \textbf{74.66}
                    & \cellcolor{LightCyan} \textbf{74.08}         & \cellcolor{LightCyan} \textbf{73.62}             & \cellcolor{LightCyan} \textbf{74.89}             & \cellcolor{LightCyan} \textbf{75.34}
                    & \cellcolor{LightCyan} \textbf{74.20}         & \cellcolor{LightCyan} \textbf{73.85}             & \cellcolor{LightCyan} \textbf{75.06}             & \cellcolor{LightCyan} \textbf{74.71}\\
\cmidrule(lr){2-12} \cmidrule(lr){14-24}
& \multirow{4}{*}{DPP} 
& Na\"ive ICL         & \textbf{18.33}
                    & 16.07         & 11.00             & \phantom{0}8.80   & \phantom{0}5.47
                    & 15.73         & 14.07             & 11.27             & \phantom{0}9.87 
&
& \multirow{4}{*}{DPP} 
& Na\"ive ICL         & 74.13 
                    & 66.84         & 52.70             & 36.78             & 20.17 
                    & 71.67         & 66.38             & 61.72             & 51.38  \\
&  
& Random     & - 
                    & 14.87         & 11.93             & \phantom{0}7.80   & \phantom{0}5.00
                    & 15.87         & 13.73             & 10.93             & \phantom{0}8.87 
&&  
& Random     & -
                    & 67.70         & 56.09             & 38.85             & 21.72
                    & 72.47         & 68.39             & 61.61             & 55.34 \\
&
& LPR               & 13.07
                    & 13.40         & 12.47             & 10.47             & \phantom{0}7.20
                    & 12.67         & 11.13             & \phantom{0}9.47   & \phantom{0}7.40 
&&
& LPR               & 67.64
                    & 65.29         & 56.78             & 43.68             & 24.83 
                    & 65.75         & 62.76             & 54.48             & 44.94 \\
&
& \cellcolor{LightCyan} Ours     & \cellcolor{LightCyan} 16.93
                    & \cellcolor{LightCyan} \textbf{16.13}         & \cellcolor{LightCyan} \textbf{16.40}             & \cellcolor{LightCyan} \textbf{16.60}             & \cellcolor{LightCyan} \textbf{14.80} 
                    & \cellcolor{LightCyan} \textbf{15.93}         & \cellcolor{LightCyan} \textbf{16.47}             & \cellcolor{LightCyan} \textbf{16.20}             & \cellcolor{LightCyan} \textbf{14.67} 
&&
& \cellcolor{LightCyan} Ours     & \cellcolor{LightCyan} \textbf{74.94} 
                    & \cellcolor{LightCyan} \textbf{74.25}         & \cellcolor{LightCyan} \textbf{74.31}             & \cellcolor{LightCyan} \textbf{74.71}             & \cellcolor{LightCyan} \textbf{74.43} 
                    & \cellcolor{LightCyan} \textbf{73.97}         & \cellcolor{LightCyan} \textbf{74.43}             & \cellcolor{LightCyan} \textbf{74.37}             & \cellcolor{LightCyan} \textbf{74.83} \\
\hline

\multirow{12}{*}{WebQ}
& \multirow{4}{*}{Random} 
& Na\"ive ICL         & 22.64 
                    & 17.97         & 11.84             & \phantom{0}7.72   & \phantom{0}3.65 
                    & 19.37         & 16.52             & 12.45             & \phantom{0}9.40 
&\multirow{12}{*}{SQuAD}
& \multirow{4}{*}{Random} 
& Na\"ive ICL         & 34.70 
                    & 34.97         & 32.87             & 26.53             & 19.67 
                    & 34.27         & 32.77             & 30.93             & 29.20 \\
&  
& Random     & -
                    & 18.25         & 12.34             & \phantom{0}7.34   & 3.96 
                    & 20.50         & 15.94             & 13.30             & \phantom{0}8.93 
&&
& Random     & - 
                    & 34.93         & 31.60             & 26.67             & 18.73 
                    & 33.90         & 33.87             & 30.67             & 28.43 \\
&
& LPR               & 22.04 
                    & 17.70         & 13.82             & \phantom{0}9.12   & \phantom{0}3.82
                    & 17.64         & 14.15             & 10.58             & \phantom{0}7.53 
&&
& LPR               & 28.27 
                    & 28.03         & 27.80             & 24.67             & 19.73 
                    & 27.30         & 27.00             & 26.13             & 24.40 \\
&
& \cellcolor{LightCyan} Ours     & \cellcolor{LightCyan} \textbf{23.41} 
                    & \cellcolor{LightCyan} \textbf{23.19}         & \cellcolor{LightCyan} \textbf{23.06}             & \cellcolor{LightCyan} \textbf{23.14}             & \cellcolor{LightCyan} \textbf{23.22} 
                    & \cellcolor{LightCyan} \textbf{23.17}         & \cellcolor{LightCyan} \textbf{22.20}             & \cellcolor{LightCyan} \textbf{21.13}             & \cellcolor{LightCyan} \textbf{19.62} 
&&
& \cellcolor{LightCyan} Ours     & \cellcolor{LightCyan} \textbf{35.17} 
                    & \cellcolor{LightCyan} \textbf{34.77}         & \cellcolor{LightCyan} \textbf{35.53}             & \cellcolor{LightCyan} \textbf{35.57}             & \cellcolor{LightCyan} \textbf{35.20} 
                    & \cellcolor{LightCyan} \textbf{34.83}         & \cellcolor{LightCyan} \textbf{35.73}             & \cellcolor{LightCyan} \textbf{35.67}             & \cellcolor{LightCyan} \textbf{35.20}\\
\cmidrule(lr){2-12} \cmidrule(lr){14-24}
& \multirow{4}{*}{TopK} 
& Na\"ive ICL         & \textbf{44.11} 
                    & 34.98         & 24.92             & 15.77             & \phantom{0}7.09 
                    & 36.55         & 29.60             & 21.19             & 14.48 
&
& \multirow{4}{*}{TopK} 
& Na\"ive ICL         & 34.47  
                    & 33.40         & 30.53             & 24.63             & 17.63
                    & 33.47         & 31.47             & 29.37             & 26.03 \\
&  
& Random     & - 
                    & 32.89         & 20.47             & 11.95             & \phantom{0}5.94 
                    & 34.73         & 26.38             & 18.71             & 12.39 
&&  
& Random     & -
                    & 32.97         & 30.37             & 26.10             & 17.43
                    & 33.53         & 31.97             & 30.87             & 26.63\\
&
& LPR               & 26.49 
                    & 22.81         & 19.48             & 14.37             & \phantom{0}6.35 
                    & 21.19         & 18.33             & 14.10             & \phantom{0}9.59 
&&
& LPR               & 34.87
                    & 26.13         & 25.27             & 21.93             & 17.67
                    & 25.50         & 25.23             & 24.00             & 21.93\\
&
& \cellcolor{LightCyan} Ours     & \cellcolor{LightCyan} 36.33 
                    & \cellcolor{LightCyan} \textbf{37.76}         & \cellcolor{LightCyan} \textbf{35.72}             & \cellcolor{LightCyan} \textbf{34.13}             & \cellcolor{LightCyan} \textbf{29.29} 
                    & \cellcolor{LightCyan} \textbf{36.85}         & \cellcolor{LightCyan} \textbf{33.17}             & \cellcolor{LightCyan} \textbf{30.92}             & \cellcolor{LightCyan} \textbf{26.30} 
&&
& \cellcolor{LightCyan} Ours     & \cellcolor{LightCyan} \textbf{35.33}
                    & \cellcolor{LightCyan} \textbf{35.57}         & \cellcolor{LightCyan} \textbf{35.73}             & \cellcolor{LightCyan} \textbf{36.80}             & \cellcolor{LightCyan} \textbf{36.37}
                    & \cellcolor{LightCyan} \textbf{36.07}         & \cellcolor{LightCyan} \textbf{35.90}             & \cellcolor{LightCyan} \textbf{35.97}             & \cellcolor{LightCyan} \textbf{35.83}\\
\cmidrule(lr){2-12} \cmidrule(lr){14-24}
 & \multirow{4}{*}{DPP} 
& Na\"ive ICL         & \textbf{45.12}
                    & 36.60         & 25.94             & 16.21             & \phantom{0}7.53 
                    & 37.54         & 30.01             & 21.60             & 14.81 
&& \multirow{4}{*}{DPP} 
& Na\"ive ICL         & 36.47 
                    & 35.23         & 31.83             & 25.93             & 17.73 
                    & 35.87         & 33.87             & 30.93             & 27.37\\
&  
& Random     & - 
                    & 34.76         & 21.74             & 11.62             & \phantom{0}5.99
                    & 36.33         & 26.94             & 17.61             & 11.54 
&&  
& Random     & - 
                    & 34.90         & 31.73             & 25.73             & 17.77 
                    & 35.40         & 33.97             & 30.63             & 27.27\\
&
& LPR               & 26.68 
                    & 22.23         & 18.88             & 12.97             & \phantom{0}6.10 
                    & 21.52         & 17.78             & 14.21             & \phantom{0}8.93 
&&
& LPR               & 26.03 
                    & 26.33         & 25.97             & 24.23             & 18.67 
                    & 25.77         & 25.53             & 24.47             & 22.23 \\
&
& \cellcolor{LightCyan} Ours     & \cellcolor{LightCyan} 37.04 
                    & \cellcolor{LightCyan} \textbf{38.09}         & \cellcolor{LightCyan} \textbf{35.94}             & \cellcolor{LightCyan} \textbf{34.10}             & \cellcolor{LightCyan} \textbf{29.57} 
                    & \cellcolor{LightCyan} \textbf{37.65}         & \cellcolor{LightCyan} \textbf{33.33}             & \cellcolor{LightCyan} \textbf{31.30}             & \cellcolor{LightCyan} \textbf{26.22} 
&&
& \cellcolor{LightCyan} Ours     & \cellcolor{LightCyan} \textbf{37.20} 
                    & \cellcolor{LightCyan} \textbf{36.53}         & \cellcolor{LightCyan} \textbf{36.70}             & \cellcolor{LightCyan} \textbf{37.10}             & \cellcolor{LightCyan} \textbf{36.87} 
                    & \cellcolor{LightCyan} \textbf{37.40}         & \cellcolor{LightCyan} \textbf{37.37}             & \cellcolor{LightCyan} \textbf{37.27}             & \cellcolor{LightCyan} \textbf{36.30}\\

\thickhline
\end{tabular}}
\caption{In various datasets, we compare four algorithms under two types of noise (relevant and irrelevant) and three retriever settings. The reported results represent the average performance across three different Random seed, and the best performing cases are highlighted in \textbf{bold}.}
\label{tab:main}
\end{table*}

\subsection{Main results}
As shown in~\autoref{tab:main}, we see a substantial improvement in the performance of the proposed algorithm across all degrees of noise and all retriever types. 
In particular, when the noise level is high, \ie $0.8$, our algorithm outperforms the naive ICL approach (\ie without any robust ICL method) by the largest margin. 
For example, on the SCIQ dataset with $0.8$ irrelevant noise, using the TopK retriever improves performance by $37.94$. 
Moreover, under the same setting for relevant noise, we observe an increase of $26.49$. 
These results indicate that the proposed algorithm can operate robustly under noisy ICL.

\subsection{Analysis}

\myparagraph{Different LLMs for noise detection.} We utilize three other distinct LLMs to calculate the $\{\mathcal{I}_{i}\}_{i=1}^N$ for noise detection  and assess the final ICL performance. 
We employ \code{GPT-Neo-1.3B}~\citep{gpt-neo} as a representative of smaller and weaker LLMs, \code{Gemma-2b}~\citep{team2024gemma} as a smaller yet potent LLM, and \code{Mistral-7B-v0.1}~\citep{jiang2023mistral} to represent LLMs of comparable size with strong capabilities, compared with our default metric model \code{Llama-2-7B}~\citep{touvron2023llama}. 
As illustrated in~\autoref{tab:diff-metric-model}, even a relatively small and less capable LLM such as \code{GPT-Neo-1.3B} exhibits only a negligible performance decline in both relevant and irrelevant noise settings, even with a high noise ratio of $0.6$.
This shows the robustness of our method, even when applied using smaller, less capable LLMs.
Additionally, the stability of our method with smaller LLMs suggests that it can be generalized to scenarios with severe computational constraints, employing smaller LLMs to compute $\{\mathcal{I}_{i}\}_{i=1}^N$. 
We also compare the AUC gain for noise detection by applying our dual-debiasing method on different LLMs as shown in~\autoref{tab:app_llm_auc_delta} of~\autoref{app:llm_choice}, which further reveals that our method is particularly
valuable for resource-constrained scenarios.

\begin{table}[H]
\centering
\resizebox{.95\columnwidth}{!}{
\begin{tabular}{lllcccc}
\thickhline
\multirow{2}{*}{\textbf{Dataset}} & \multirow{2}{*}{\textbf{Retriever}} & \multirow{2}{*}{\textbf{Metric Model}} 
& \multicolumn{2}{c}{\textbf{Irrelevant Noise}} 
& \multicolumn{2}{c}{\textbf{Relevant Noise}} \\
\cmidrule(lr){4-5} \cmidrule(lr){6-7}
& &
& \textbf{0.4} & \textbf{0.6}  
& \textbf{0.4} & \textbf{0.6}  \\
\midrule

\multirow{12}{*}{NQ}
 & \multirow{4}{*}{Random} 
& \code{GPT-Neo-1.3B}        
                   & 13.87    &   13.10    &   14.13    &   12.87  \\
&  
& \code{Gemma-2b}   
                   & 13.93    &   13.84    &   13.80    &   13.00 \\
&
& \code{Mistral-7B-v0.1}              
                   & 13.80    &   13.37    &   13.87    &   13.57 \\
&
& \code{Llama-2-7B}      
                   & 13.93    &   13.67    &   13.33    &   13.73 \\
\cmidrule(lr){2-7}
 & \multirow{4}{*}{TopK} 
& \code{GPT-Neo-1.3B}         
                   & 15.47    &   15.00    &   14.07    &   13.80 \\
&  
& \code{Gemma-2b}     
                    & 16.20    &   16.20    &   16.00    &   15.77 \\
&
& \code{Mistral-7B-v0.1}               
                    & 16.00    &   16.54    &   16.47    &   16.14 \\
&
& \code{Llama-2-7B}       
                    & 16.33    &   15.70    &   15.87    &   15.64 \\
\cmidrule(lr){2-7}
& \multirow{4}{*}{DPP} 
& \code{GPT-Neo-1.3B}          
                    & 16.53    &   15.54    &   14.07    &   13.84 \\
&  
& \code{Gemma-2b}      
                    & 16.33    &   16.00    &   16.13    &   14.73 \\
&
& \code{Mistral-7B-v0.1}              
                    & 16.20    &   15.80    &   16.47    &   15.40 \\
&
& \code{Llama-2-7B}     
                    & 16.40    &   16.34    &   16.47    &   15.74 \\
\hline
 
\multirow{12}{*}{SciQ}
 & \multirow{4}{*}{Random} 
& \code{GPT-Neo-1.3B}        
                    & 75.69    &   75.29    &   75.09    &   74.71 \\
&  
& \code{Gemma-2b}      
                    & 76.04    &   75.69    &   76.04    &   75.69 \\
&
& \code{Mistral-7B-v0.1}                
                    & 76.18    &   76.01    &   75.58    &   75.98 \\
&
& \code{Llama-2-7B}      
                    & 75.84    &   75.43    &   75.98    &   76.41 \\
\cmidrule(lr){2-7}
 & \multirow{4}{*}{TopK} 
& \code{GPT-Neo-1.3B}         
                    & 73.45    &   75.00    &   74.43    &   74.00 \\
&  
& \code{Gemma-2b}      
                    & 73.42    &   74.57    &   74.17    &   74.34 \\
&
& \code{Mistral-7B-v0.1}               
                    & 73.60    &   75.37    &   74.20    &   75.43 \\
&
& \code{Llama-2-7B}       
                    & 73.57    &   75.00    &   73.82    &   75.18 \\
\cmidrule(lr){2-7}
 & \multirow{4}{*}{DPP} 
& \code{GPT-Neo-1.3B}          
                    & 74.25    &   74.20    &   75.40    &   74.20 \\
&  
& \code{Gemma-2b}      
                    & 73.97    &   74.66    &   74.37    &   73.22 \\
&
& \code{Mistral-7B-v0.1}                
                    & 74.37    &   75.03    &   74.31    &   74.83 \\
&
& \code{Llama-2-7B}      
                    & 74.31    &   74.83    &   74.43    &   74.37 \\
\thickhline
\end{tabular}}
\caption{ICL performance of using different LLM as the metric model. The default inference LLM is \code{Llama-2-7B}.}
\label{tab:diff-metric-model}
\end{table}

\begin{figure}[H]
        \centering
        \includegraphics[width=1.0\columnwidth]{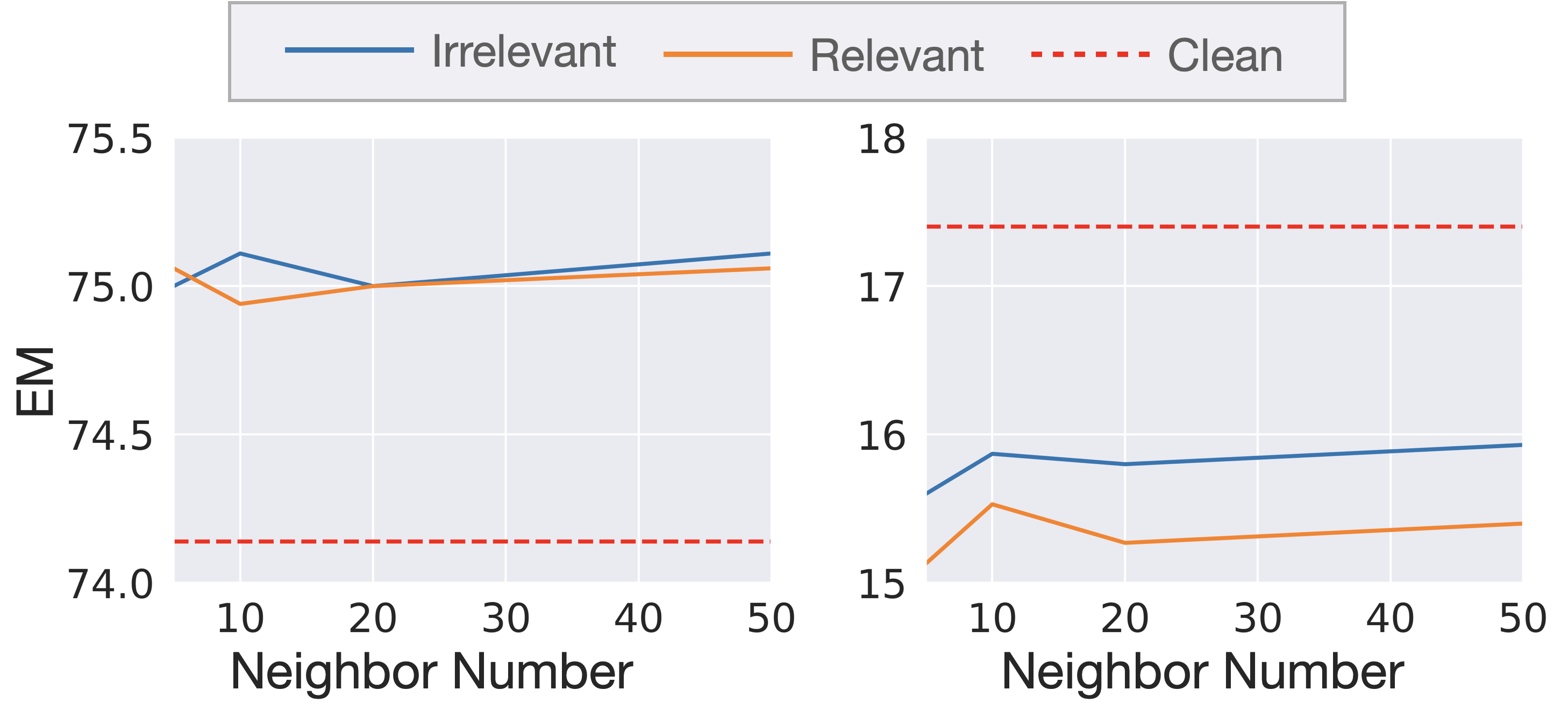}    
        \caption{Sensitivity on neighbor number $N_{\text{neighbor}}$ under noise ratio $0.6$ and TopK retriever. Left: SCIQ; right: NQ. The `irrelevant'/`relevant' indicates the performance of our method under noise, and `clean' indicates the performance of naive baseline under clean.}
        \label{fig:comp-efficient-neighbor-num}
\end{figure}



\myparagraph{Computation cost.} 
We analyze the proposed algorithm based on two key factors. 
First, we compare its performance when using different metric models. 
If a smaller model achieves performance comparable to that of a larger model, it indicates that the approach can be utilized efficiently. 
As shown in~\autoref{tab:diff-metric-model}, the performance remains similar regardless of whether a \code{1.3B} or a \code{7B} model is used. 
This suggests that employing a smaller model as the metric model does not lead to significant performance degradation, enabling more efficient utilization. 
Second, we examine the impact of controlling the neighbor size $N_{\text{neighbor}}$, as illustrated in~\autoref{fig:comp-efficient-neighbor-num}. 
The results show that the performance remains consistent even as the number of neighbors increases. 
This suggests that the proposed algorithm operates efficiently even with relatively few neighbors. 
Based on these findings, we highlight that the proposed algorithm is both efficient and robust.



\begin{figure}[t!]
        \centering
        \includegraphics[width=1.0\columnwidth]{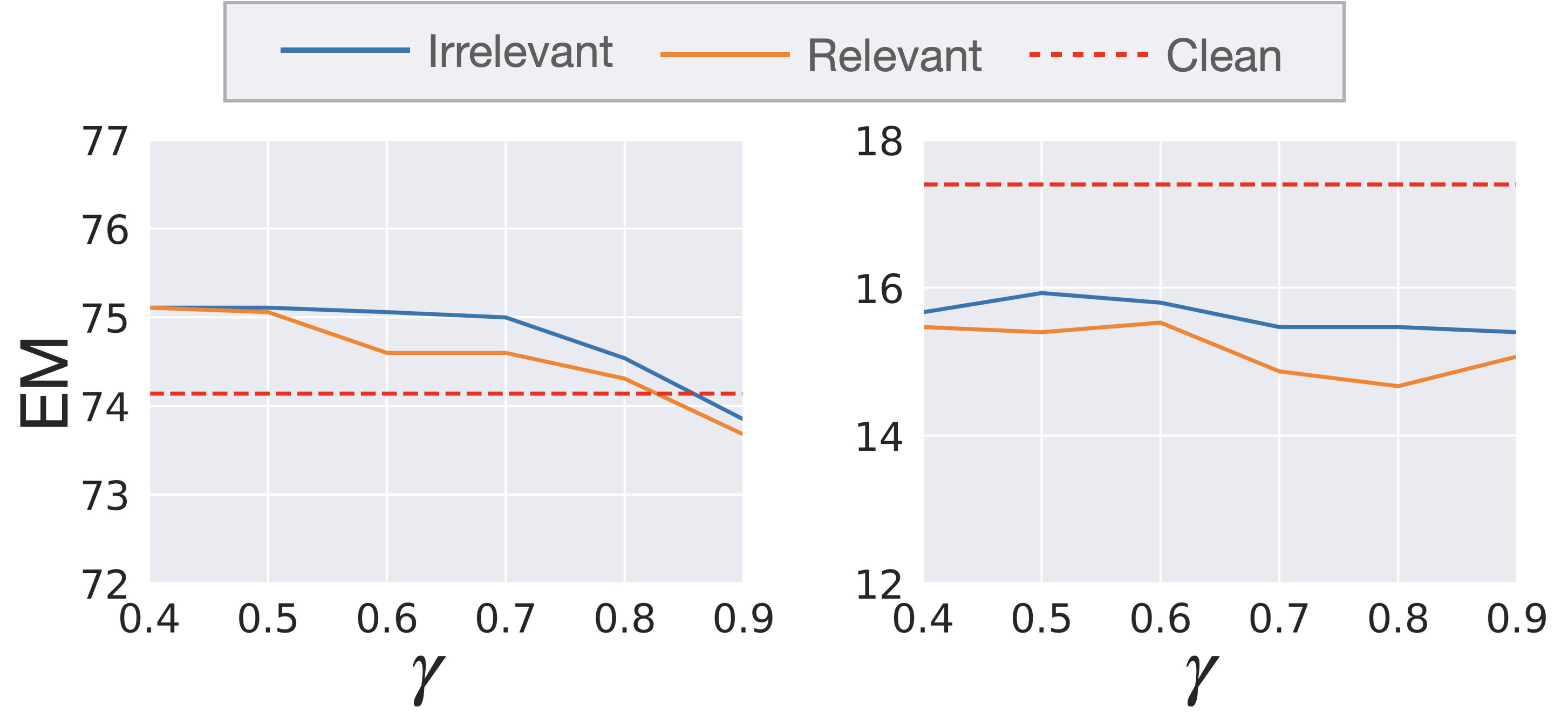}    
        \caption{Sensitivity of $\gamma$ under noise ratio $0.6$ and TopK retriever. Left: SCIQ; right: NQ. The `irrelevant'/`relevant' indicates the performance of our method under noised setting, and `clean' indicates the performance of naive ICL baseline under clean setting.}
        \label{fig:sensitivity_gamma}
\end{figure}

\myparagraph{Sensitivity  of $\gamma$:} We examine our method's sensitivity on the probability threshold $\gamma$ involved in noise detection under the setting of noise ratio $0.6$ using TopK retriever. As shown in~\autoref{fig:sensitivity_gamma}, the performance of our method remains consistent on both NQ and SCIQ datasets when $\gamma \in [0.4, 0.9]$, with negligible drop from naive ICL of the clean setting, even the majority of the training set is noised. 
This shows that our method stays robust on different $\gamma$, and the default $\gamma=0.5$ can be a proper choice for various noise settings.



\myparagraph{Qualitative analysis of failure cases.}
We conducted an in-depth analysis of failure cases on WebQ with $40\%$ relevant noise, revealing a critical pattern: \textit{annotation length significantly impacts detection accuracy}. 
We present the annotation sequence length distribution for clean/noisy samples that our method failed/successfully detected in~\autoref{fig:app_failed_cases_seq_length} of~\autoref{app:failure_case}, respectively. Specifically: \quad 1) Successfully detected clean samples: Only $14\%$ had annotation lengths larger than 7 tokens; \quad 2) Successfully detected noisy samples: Only $12\%$ had annotation lengths larger than 7 tokens; \quad 3) Failed detection cases (noisy misclassified as clean): $36\%$ had annotation lengths larger than 7 tokens. \quad
This pattern is theoretically consistent with our formulation on the estimation of extrinsic bias, as longer annotations create more complex probability distributions that can obscure the distinction between clean and noisy samples. 
This finding provides actionable insights for future work: adaptive debiasing techniques could be designed to account for annotation length, potentially using length-normalized probability estimates or employing specialized models for longer sequences.

\section{Related Work}
\label{sec:related}

\myparagraph{In-context learning (ICL):} Recent research has leveraged pre-trained LLMs for downstream NLP tasks through in-context learning, particularly in text classification~\citep{yoo2022ground} and generation tasks~\citep{o2023contrastive}. Notable advances include the UDR retriever by \citet{li2023unified}, which works effectively across multiple tasks, and the efficient approach by \citet{liucontext} that extracts in-context vectors from LLM embeddings to reduce computational costs. However, most ICL research assumes clean, high-quality demonstrations, leaving open questions about performance with noisy or imperfect examples.

\myparagraph{ICL with noisy annotations:} Initial studies exploring random labels in ICL classification have shown mixed results. While \citet{min2022rethinking} found limited performance impact with random retrievers for certain LLM-dataset combinations, \citet{yoo2022ground} demonstrated significant performance degradation across a broader range of settings.
More recent work has begun addressing noisy ICL directly. \citet{kang2024context} proposed \textit{Rectification} for classification tasks, though its fine-tuning requirements introduce substantial computational overhead. For generation tasks, \citet{gao2024noise} pioneered the first noise-robust method, but it shows limitations under high-noise conditions.

\myparagraph{Debiasing LLM Output:} Despite their capabilities, LLMs can exhibit biases from their pre-training corpora that impact task performance. To address this, \citet{li2022contrastive} and \citet{zhao2024enhancing} developed Contrastive Decoding, which improves text generation quality by debiasing larger LLMs using outputs from smaller models within the same family. Additionally, \citet{fei2023mitigating} and \citet{zhao2021calibrate} introduced methods to reduce bias in LLMs by addressing both prefixed context bias and finite label bias in classification tasks.
\section{Conclusion}
\label{sec:conclusion}
In this paper, we have presented a robust method for handling noisy demonstrations in In-Context Learning (ICL). 
Our approach addresses both intrinsic and extrinsic biases in LLMs' perception of noisy demonstrations, enabling more reliable detection of problematic examples and improving the overall inference process. Through extensive experiments across four text generation datasets, we demonstrate our method's effectiveness under various retrieval strategies and noise conditions, particularly showing strong performance even with noise ratios as high as 0.8. 
Notably, our approach achieves competitive performance without significant computational overhead, making it practical for real-world applications. 
We also show that our method maintains its effectiveness even when using smaller LLMs for metric computation, further enhancing its practical utility. 
The successful results across different settings validate not only the theoretical foundations of our dual debiasing framework but also its practical applicability. 
Through this work, we advance both the robust deployment of LLMs and our understanding of how these models perceive and process noisy query-annotation pairs, providing a foundation for future research in robust ICL methods.



\section*{Limitations}
While our dual debiasing approach demonstrates strong performance across multiple datasets and LLM architectures, several limitations should be acknowledged. First, our method's effectiveness relies heavily on the quality and diversity of the neighbor samples used for extrinsic debiasing. In domains with limited available data or highly specialized knowledge, generating appropriate neighbors may be challenging, potentially affecting the accuracy of our {\it Sample Cleanliness Score.} Additionally, while our approach works well for text generation tasks, its applicability to other ICL applications like classification or structured prediction remains unexplored.

\section*{Potential Risks}
The computational cost of neighbor generation and metric calculation, though moderate, increases linearly with the number of demonstration samples. While we show that using fewer neighbors can maintain performance, there may be a practical upper limit to the dataset size our method can handle efficiently. Furthermore, there is a risk that our approach might inadvertently discard valid but unusual demonstrations that appear noisy due to their uniqueness, particularly in specialized domains where LLMs have limited exposure during pre-training. Users should carefully consider these limitations when applying our method to new domains or tasks.

\section*{Acknowledgments}
This material is based in part upon work supported by the National Science Foundation under Grant IIS-2212174, National Institute of Aging (NIA) 1RF1AG072449, National Institute of General Medical Sciences (NIGMS) 1R01GM145700.


\bibliography{references}

\begin{thebibliography}{29}
\providecommand{\natexlab}[1]{#1}

\bibitem[{Achiam et~al.(2023)Achiam, Adler, Agarwal, Ahmad, Akkaya, Aleman, Almeida, Altenschmidt, Altman, Anadkat et~al.}]{achiam2023gpt}
Josh Achiam, Steven Adler, Sandhini Agarwal, Lama Ahmad, Ilge Akkaya, Florencia~Leoni Aleman, Diogo Almeida, Janko Altenschmidt, Sam Altman, Shyamal Anadkat, et~al. 2023.
\newblock Gpt-4 technical report.
\newblock \emph{arXiv preprint arXiv:2303.08774}.

\bibitem[{Alon and Kamfonas(2023)}]{alon2023detecting}
Gabriel Alon and Michael Kamfonas. 2023.
\newblock Detecting language model attacks with perplexity.
\newblock \emph{arXiv preprint arXiv:2308.14132}.

\bibitem[{Arora et~al.(2021)Arora, Huang, and He}]{arora2021types}
Udit Arora, William Huang, and He~He. 2021.
\newblock Types of out-of-distribution texts and how to detect them.
\newblock \emph{arXiv preprint arXiv:2109.06827}.

\bibitem[{Black et~al.(2021)Black, Leo, Wang, Leahy, and Biderman}]{gpt-neo}
Sid Black, Gao Leo, Phil Wang, Connor Leahy, and Stella Biderman. 2021.
\newblock \href {https://doi.org/10.5281/zenodo.5297715} {{GPT-Neo: Large Scale Autoregressive Language Modeling with Mesh-Tensorflow}}.
\newblock {If you use this software, please cite it using these metadata.}

\bibitem[{Brown et~al.(2020)Brown, Mann, Ryder, Subbiah, Kaplan, Dhariwal, Neelakantan, Shyam, Sastry, Askell et~al.}]{brown2020language}
Tom Brown, Benjamin Mann, Nick Ryder, Melanie Subbiah, Jared~D Kaplan, Prafulla Dhariwal, Arvind Neelakantan, Pranav Shyam, Girish Sastry, Amanda Askell, et~al. 2020.
\newblock Language models are few-shot learners.
\newblock \emph{Advances in neural information processing systems}, 33:1877--1901.

\bibitem[{Dong et~al.(2024)Dong, Li, Dai, Zheng, Ma, Li, Xia, Xu, Wu, Chang, Sun, Li, and Sui}]{dong-etal-2024-survey}
Qingxiu Dong, Lei Li, Damai Dai, Ce~Zheng, Jingyuan Ma, Rui Li, Heming Xia, Jingjing Xu, Zhiyong Wu, Baobao Chang, Xu~Sun, Lei Li, and Zhifang Sui. 2024.
\newblock \href {https://doi.org/10.18653/v1/2024.emnlp-main.64} {A survey on in-context learning}.
\newblock In \emph{Proceedings of the 2024 Conference on Empirical Methods in Natural Language Processing}, pages 1107--1128, Miami, Florida, USA. Association for Computational Linguistics.

\bibitem[{Fei et~al.(2023)Fei, Hou, Chen, and Bosselut}]{fei2023mitigating}
Yu~Fei, Yifan Hou, Zeming Chen, and Antoine Bosselut. 2023.
\newblock Mitigating label biases for in-context learning.
\newblock \emph{arXiv preprint arXiv:2305.19148}.

\bibitem[{Gao et~al.(2024)Gao, Zhang, Jiang, Shu, Zheng, and Wei}]{gao2024noise}
Hongfu Gao, Feipeng Zhang, Wenyu Jiang, Jun Shu, Feng Zheng, and Hongxin Wei. 2024.
\newblock On the noise robustness of in-context learning for text generation.

\bibitem[{Huang et~al.(2023)Huang, Zhang, and Shan}]{huang2023twin}
Zhizhong Huang, Junping Zhang, and Hongming Shan. 2023.
\newblock Twin contrastive learning with noisy labels.
\newblock In \emph{Proceedings of the IEEE/CVF Conference on Computer Vision and Pattern Recognition}, pages 11661--11670.

\bibitem[{Jiang et~al.(2023)Jiang, Sablayrolles, Mensch, Bamford, Chaplot, Casas, Bressand, Lengyel, Lample, Saulnier et~al.}]{jiang2023mistral}
Albert~Q Jiang, Alexandre Sablayrolles, Arthur Mensch, Chris Bamford, Devendra~Singh Chaplot, Diego de~las Casas, Florian Bressand, Gianna Lengyel, Guillaume Lample, Lucile Saulnier, et~al. 2023.
\newblock Mistral 7b.
\newblock \emph{arXiv preprint arXiv:2310.06825}.

\bibitem[{Kang et~al.(2024)Kang, Son, Song, and Chang}]{kang2024context}
Junyong Kang, Donghyun Son, Hwanjun Song, and Buru Chang. 2024.
\newblock In-context learning with noisy labels.
\newblock \emph{arXiv preprint arXiv:2411.19581}.

\bibitem[{Karim et~al.(2022)Karim, Rizve, Rahnavard, Mian, and Shah}]{karim2022unicon}
Nazmul Karim, Mamshad~Nayeem Rizve, Nazanin Rahnavard, Ajmal Mian, and Mubarak Shah. 2022.
\newblock Unicon: Combating label noise through uniform selection and contrastive learning.
\newblock In \emph{Proceedings of the IEEE/CVF conference on computer vision and pattern recognition}, pages 9676--9686.

\bibitem[{Kwiatkowski et~al.(2019)Kwiatkowski, Palomaki, Redfield, Collins, Parikh, Alberti, Epstein, Polosukhin, Devlin, Lee et~al.}]{kwiatkowski2019natural}
Tom Kwiatkowski, Jennimaria Palomaki, Olivia Redfield, Michael Collins, Ankur Parikh, Chris Alberti, Danielle Epstein, Illia Polosukhin, Jacob Devlin, Kenton Lee, et~al. 2019.
\newblock Natural questions: a benchmark for question answering research.
\newblock \emph{Transactions of the Association for Computational Linguistics}, 7:453--466.

\bibitem[{Li et~al.()Li, Socher, and Hoi}]{lidividemix}
Junnan Li, Richard Socher, and Steven~CH Hoi.
\newblock Dividemix: Learning with noisy labels as semi-supervised learning.
\newblock In \emph{International Conference on Learning Representations}.

\bibitem[{Li et~al.(2022)Li, Holtzman, Fried, Liang, Eisner, Hashimoto, Zettlemoyer, and Lewis}]{li2022contrastive}
Xiang~Lisa Li, Ari Holtzman, Daniel Fried, Percy Liang, Jason Eisner, Tatsunori Hashimoto, Luke Zettlemoyer, and Mike Lewis. 2022.
\newblock Contrastive decoding: Open-ended text generation as optimization.
\newblock \emph{arXiv preprint arXiv:2210.15097}.

\bibitem[{Li et~al.(2023)Li, Lv, Yan, Lin, Zhu, Ni, Xie, Wang, and Qiu}]{li2023unified}
Xiaonan Li, Kai Lv, Hang Yan, Tianyang Lin, Wei Zhu, Yuan Ni, Guotong Xie, Xiaoling Wang, and Xipeng Qiu. 2023.
\newblock Unified demonstration retriever for in-context learning.
\newblock In \emph{Proceedings of the 61st Annual Meeting of the Association for Computational Linguistics (Volume 1: Long Papers)}, pages 4644--4668.

\bibitem[{Liu et~al.()Liu, Ye, Xing, and Zou}]{liucontext}
Sheng Liu, Haotian Ye, Lei Xing, and James~Y Zou.
\newblock In-context vectors: Making in context learning more effective and controllable through latent space steering.
\newblock In \emph{Forty-first International Conference on Machine Learning}.

\bibitem[{Min et~al.(2022)Min, Lyu, Holtzman, Artetxe, Lewis, Hajishirzi, and Zettlemoyer}]{min2022rethinking}
Sewon Min, Xinxi Lyu, Ari Holtzman, Mikel Artetxe, Mike Lewis, Hannaneh Hajishirzi, and Luke Zettlemoyer. 2022.
\newblock Rethinking the role of demonstrations: What makes in-context learning work?
\newblock \emph{arXiv preprint arXiv:2202.12837}.

\bibitem[{O'Brien and Lewis(2023)}]{o2023contrastive}
Sean O'Brien and Mike Lewis. 2023.
\newblock Contrastive decoding improves reasoning in large language models.
\newblock \emph{arXiv preprint arXiv:2309.09117}.

\bibitem[{Radford et~al.(2019)Radford, Wu, Child, Luan, Amodei, Sutskever et~al.}]{radford2019language}
Alec Radford, Jeffrey Wu, Rewon Child, David Luan, Dario Amodei, Ilya Sutskever, et~al. 2019.
\newblock Language models are unsupervised multitask learners.
\newblock \emph{OpenAI blog}, 1(8):9.

\bibitem[{Team et~al.(2024)Team, Riviere, Pathak, Sessa, Hardin, Bhupatiraju, Hussenot, Mesnard, Shahriari, Ram{\'e} et~al.}]{team2024gemma}
Gemma Team, Morgane Riviere, Shreya Pathak, Pier~Giuseppe Sessa, Cassidy Hardin, Surya Bhupatiraju, L{\'e}onard Hussenot, Thomas Mesnard, Bobak Shahriari, Alexandre Ram{\'e}, et~al. 2024.
\newblock Gemma 2: Improving open language models at a practical size.
\newblock \emph{arXiv preprint arXiv:2408.00118}.

\bibitem[{Touvron et~al.(2023)Touvron, Lavril, Izacard, Martinet, Lachaux, Lacroix, Rozi{\`e}re, Goyal, Hambro, Azhar et~al.}]{touvron2023llama}
Hugo Touvron, Thibaut Lavril, Gautier Izacard, Xavier Martinet, Marie-Anne Lachaux, Timoth{\'e}e Lacroix, Baptiste Rozi{\`e}re, Naman Goyal, Eric Hambro, Faisal Azhar, et~al. 2023.
\newblock Llama: Open and efficient foundation language models.
\newblock \emph{arXiv preprint arXiv:2302.13971}.

\bibitem[{Wei et~al.(2022)Wei, Wang, Schuurmans, Bosma, Xia, Chi, Le, Zhou et~al.}]{wei2022chain}
Jason Wei, Xuezhi Wang, Dale Schuurmans, Maarten Bosma, Fei Xia, Ed~Chi, Quoc~V Le, Denny Zhou, et~al. 2022.
\newblock Chain-of-thought prompting elicits reasoning in large language models.
\newblock \emph{Advances in neural information processing systems}, 35:24824--24837.

\bibitem[{Wu et~al.(2023)Wu, Wang, Ye, Feng, Xu, Qiao, and Wu}]{wu2023openicl}
Zhenyu~Wu Wu, Yaoxiang Wang, Jiacheng Ye, Jiangtao Feng, Jingjing Xu, Yu~Qiao, and Zhiyong Wu. 2023.
\newblock Openicl: An open-source framework for in-context learning.
\newblock \emph{arXiv preprint arXiv:2303.02913}.

\bibitem[{Yan et~al.(2024)Yan, Yan, Chen, Li, Zhu, and Zhang}]{yan2024benchmarking}
Jianhao Yan, Pingchuan Yan, Yulong Chen, Jing Li, Xianchao Zhu, and Yue Zhang. 2024.
\newblock Benchmarking gpt-4 against human translators: A comprehensive evaluation across languages, domains, and expertise levels.
\newblock \emph{arXiv preprint arXiv:2411.13775}.

\bibitem[{Ye et~al.(2023)Ye, Wu, Feng, Yu, and Kong}]{ye2023compositional}
Jiacheng Ye, Zhiyong Wu, Jiangtao Feng, Tao Yu, and Lingpeng Kong. 2023.
\newblock Compositional exemplars for in-context learning.
\newblock In \emph{International Conference on Machine Learning}, pages 39818--39833. PMLR.

\bibitem[{Yoo et~al.(2022)Yoo, Kim, Kim, Cho, Jo, Lee, Lee, and Kim}]{yoo2022ground}
Kang~Min Yoo, Junyeob Kim, Hyuhng~Joon Kim, Hyunsoo Cho, Hwiyeol Jo, Sang-Woo Lee, Sang-goo Lee, and Taeuk Kim. 2022.
\newblock Ground-truth labels matter: A deeper look into input-label demonstrations.
\newblock In \emph{Proceedings of the 2022 Conference on Empirical Methods in Natural Language Processing}, pages 2422--2437.

\bibitem[{Zhao et~al.(2024)Zhao, Monti, Lehmann, and Assem}]{zhao2024enhancing}
Zheng Zhao, Emilio Monti, Jens Lehmann, and Haytham Assem. 2024.
\newblock Enhancing contextual understanding in large language models through contrastive decoding.
\newblock \emph{arXiv preprint arXiv:2405.02750}.

\bibitem[{Zhao et~al.(2021)Zhao, Wallace, Feng, Klein, and Singh}]{zhao2021calibrate}
Zihao Zhao, Eric Wallace, Shi Feng, Dan Klein, and Sameer Singh. 2021.
\newblock Calibrate before use: Improving few-shot performance of language models.
\newblock In \emph{International conference on machine learning}, pages 12697--12706. PMLR.

\end{thebibliography}

\appendix


\begin{table*}[h!]
    \centering
    \resizebox{0.9\linewidth}{!}{
    \begin{tabular}{cp{1.5cm}p{18cm}}
    \thickhline
        \textbf{Dataset}     & \textbf{Setting}                               & \textbf{In-Context Demonstration}           \\ \hline
        \multirow{6}{*}[-0.5em]{NQ}  & \multirow{2}{*}{\textcolor{blue}{Clean}}       &  \textbf{Question}: how i.met your mother who is the mother?    \\
                             &                                                &  \textbf{Answer}: \textcolor{blue}{Tracy McConnell}    \\
                             \cmidrule{2-3}
                             & \multirow{2}{*}{\textcolor{red}{Irrelevant}}   &  \textbf{Question}: how i.met your mother who is the mother?    \\
                             &                                                &  \textbf{Answer}: \textcolor{red}{Moreirense F.C.}    \\
                             \cmidrule{2-3}
                             & \multirow{2}{*}{\textcolor{red}{Relevant}}     &   \textbf{Question}: how i.met your mother who is the mother?   \\
                             &                                                &  \textbf{Answer}: \textcolor{red}{Barney Stinson is the mother}    \\
        \hline
        \multirow{6}{*}[-0.5em]{WebQ}  & \multirow{2}{*}{\textcolor{blue}{Clean}}       &  \textbf{Question}: where are the nfl redskins from?   \\
                               &                                                &  \textbf{Answer}: \textcolor{blue}{Washington Redskins}     \\
                               \cmidrule{2-3}
                               & \multirow{2}{*}{\textcolor{red}{Irrelevant}}   &  \textbf{Question}: where are the nfl redskins from?   \\
                               &                                                &  \textbf{Answer}:  \textcolor{red}{the Bee Gees}    \\
                               \cmidrule{2-3}
                               & \multirow{2}{*}{\textcolor{red}{Relevant}}     &  \textbf{Question}:  where are the nfl redskins from?   \\
                               &                                                &   \textbf{Answer}:  \textcolor{red}{Los Angeles, California}    \\
        \hline
        \multirow{9}{*}[-4.2em]{SCIQ}  & \multirow{3}{*}[-1.2em]{\textcolor{blue}{Clean}}       &  \textbf{Support}: It might only take one sperm to fertilize an egg, but that sperm is not alone. Hundreds of millions of sperm can be released during sexual intercourse.    \\
                               &                                                &  \textbf{Question}: How many sperm does it take to fertilize an egg?   \\
                               &                                            &  \textbf{Answer}: \textcolor{blue}{1}       \\
                               \cmidrule{2-3}
                               & \multirow{3}{*}[-1.2em]{\textcolor{red}{Irrelevant}}   &  \textbf{Support}:  It might only take one sperm to fertilize an egg, but that sperm is not alone. Hundreds of millions of sperm can be released during sexual intercourse.   \\
                               &                                                &  \textbf{Question}: How many sperm does it take to fertilize an egg?   \\
                               &                                                &  \textbf{Answer}:  \textcolor{red}{open clusters}       \\
                               \cmidrule{2-3}
                               & \multirow{3}{*}[-1.2em]{\textcolor{red}{Relevant}}     &  \textbf{Support}: It might only take one sperm to fertilize an egg, but that sperm is not alone. Hundreds of millions of sperm can be released during sexual intercourse.    \\
                               &                                                &  \textbf{Question}: How many sperm does it take to fertilize an egg?   \\
                               &                                                &  \textbf{Answer}:   \textcolor{red}{3}   \\
        \hline
        \multirow{9}{*}[-4.2em]{SQuAD}  & \multirow{3}{*}[-1.2em]{\textcolor{blue}{Clean}}       &  \textbf{Question}: What was the name of the streaming service?  \\
                                &                                                &   \textbf{Context}:  On February 6, 2016, one day before her performance at the Super Bowl, Beyoncé released a new single exclusively on music streaming service Tidal called ``Formation''.    \\
                                &                                                &   \textbf{Answer}: \textcolor{blue}{Tidal}   \\
                                \cmidrule{2-3}
                                & \multirow{3}{*}[-1.2em]{\textcolor{red}{Irrelevant}}   & \textbf{Question}: What was the name of the streaming service?   \\
                                &                                                &  \textbf{Context}:  On February 6, 2016, one day before her performance at the Super Bowl, Beyoncé released a new single exclusively on music streaming service Tidal called ``Formation''.    \\
                                &                                                &   \textbf{Answer}: \textcolor{red}{village}   \\
                                \cmidrule{2-3}
                                & \multirow{3}{*}[-1.2em]{\textcolor{red}{Relevant}}     &  \textbf{Question}: What was the name of the streaming service?   \\
                                &                                                &  \textbf{Context}:  On February 6, 2016, one day before her performance at the Super Bowl, Beyoncé released a new single exclusively on music streaming service Tidal called ``Formation''.  \\
                                &                                                &   \textbf{Answer}: \textcolor{red}{Spotify}   \\
    \thickhline
    \end{tabular}}
    \caption{Clean/noisy demonstration examples for each dataset.}
    \label{tab:dataset_noise_example}
\end{table*}
\begin{table*}[h!]
    \centering
    \resizebox{0.8\linewidth}{!}{
    \begin{tabular}{cp{3.7cm}p{9cm}}
    \thickhline
        \textbf{Dataset}     & \textbf{Prompt}                    & \textbf{Example}           \\ \hline
        \multirow{2}{*}{NQ}          & \textbf{Question}: \code{<Question>}          & \textbf{Question}: what do the 3 dots mean in math               \\
        & \textbf{Answer}: \code{<Answer>} &  \textbf{Answer}: the therefore sign \\
        \hline
        \multirow{2}{*}{WebQ}        & \textbf{Question}: \code{<Question>}          & \textbf{Question}: what is the oregon ducks 2012 football schedule?              \\
        &  \textbf{Answer}: \code{<Answer>}  & \textbf{Answer}: University of Oregon  \\
        \hline
        \multirow{3}{*}{SCIQ}  &   \textbf{Support}: \code{<Support>}  & \textbf{Support}: Smooth muscle regulates air flow in lungs.     \\
        &  \textbf{Question}: \code{<Question>}  & \textbf{Question}:  Which kind of muscle regulates air flow in lungs?  \\
        &  \textbf{Answer}: \code{<Answer>}  & \textbf{Answer}: smooth  \\
        \hline
        \multirow{3}{*}[-3.6em]{SQuAD}   &   \textbf{Question}: \code{<Question>}  & \textbf{Question}: Who won the Super Bowl MVP?    \\
        &  \textbf{Context}: \code{<Context>}  & \textbf{Context}: The Broncos took an early lead in Super Bowl 50 and never trailed. Newton was limited by Denver's defense, which sacked him seven times and forced him into three turnovers, including a fumble which they recovered for a touchdown. Denver linebacker Von Miller was named Super Bowl MVP, recording five solo tackles, $2 \frac{1}{2}$ sacks, and two forced fumbles.  \\
        &  \textbf{Answer}: \code{<Answer>}  & \textbf{Answer}: Von Miller  \\
    \thickhline
    \end{tabular}}
    \caption{The ICL template for datasets. Placeholders (\eg \code{<Question>} and \code{<Answer>}) will be replaced by real questions or answers.}
    \label{tab:dataset_template}
\end{table*}

\section{Experiment Setting}
\label{app:exp-setting}

\begin{table}[h]
    \centering
    \resizebox{1\linewidth}{!}{
    \begin{tabular}{cccccc}
    \thickhline
        Dataset     & Task                    & Training      & Test         \\ \hline
        NQ          & Open-Domain QA          & 10,000        & 500           \\
        WebQ        & Open-Domain QA          & 1,261        & 1,213           \\
        SCIQ        & Reading Comprehension   & 6,059        & 580             \\
        SQuAD       & Reading Comprehension   & 20,000        & 1,000                  \\
    \thickhline
    \end{tabular}}
    \caption{The statistics of the datasets used.}
    \label{tab:dataset_statistics}
\end{table}
\myparagraph{Dataset} In this study, we utilize four text generation tasks. Details of the datasets are presented in~\autoref{tab:dataset_statistics}. The complete ICL template for each dataset is shown in~\autoref{tab:dataset_template}. And the noisy demonstration examples are shown in~\autoref{tab:dataset_noise_example}.

\begin{table}[H]
    \centering
    \resizebox{1\linewidth}{!}{
    \begin{tabular}{cccc}
    \thickhline
        Dataset  &  Noise Ratio   &  \texttt{gmm\_part\_thres}                  & $\gamma$       \\ \hline
        NQ       &  $[0.2, 0.8]$          & 5.0        & 0.5          \\
        WebQ     &  $[0.2, 0.8]$           & 4.0       & 0.5          \\
        SCIQ     &  $[0.2, 0.8]$           & 10.0       & 0.5          \\
        SQUAD      &  $[0.2, 0.8]$         & 12.5       & 0.5                  \\
    \thickhline
    \end{tabular}}
    \caption{The hyper-parameter setting for the main table.}
    \label{tab:main_table_hyperparam}
\end{table}
\myparagraph{Hyper-parameter} The hyper-parameter settings for main table are shown in~\autoref{tab:main_table_hyperparam}.

\myparagraph{Computation Environment} We run our experiments on NVIDIA RTX A5000 GPU. Each experiment takes less than half an hour on a single GPU. 
\section{More Experiment Results}
\label{app:more-exp-result}

\subsection{Fewer neighbors results for our method}
\begin{figure}[H]
    \centering
    \subfloat[NQ with Random\label{fig:nq_neighbor_num_random}]{
        \includegraphics[width=0.45\columnwidth]{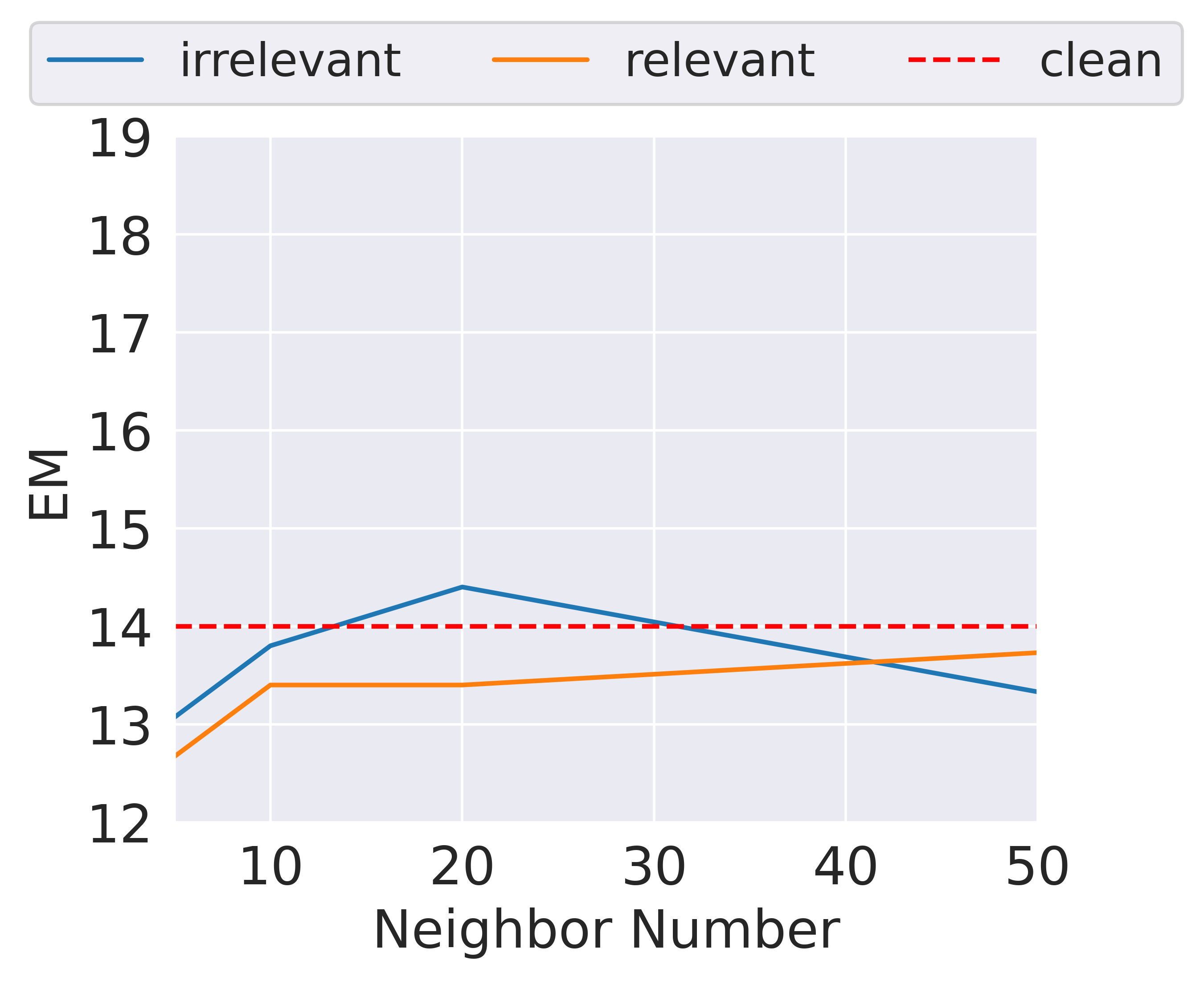}
    }
    \hspace{2mm}
    \subfloat[NQ with DPP\label{fig:nq_neighbor_num_dpp}]{
        \includegraphics[width=0.45\columnwidth]{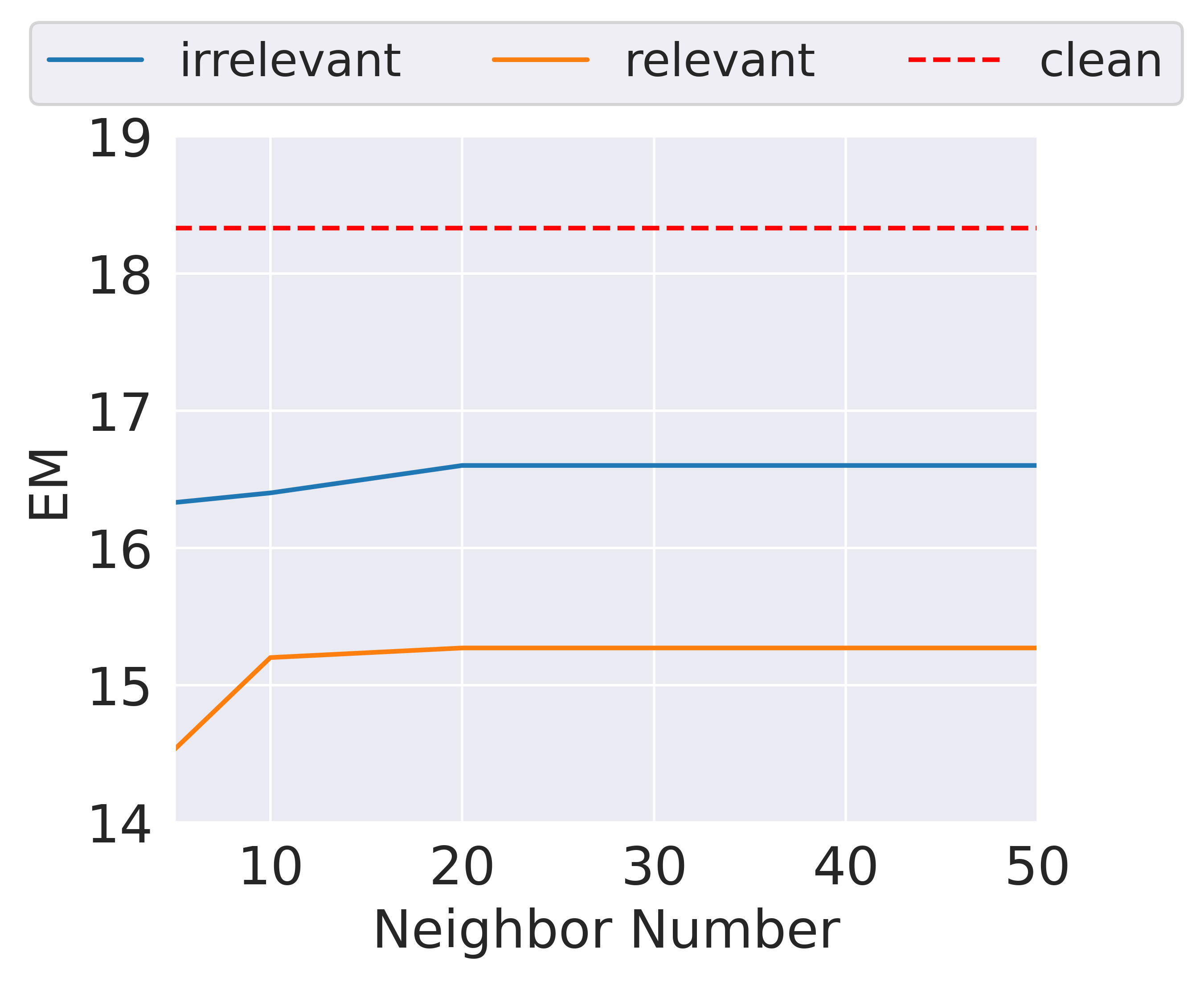}
    }
    \hfill
    \subfloat[SCIQ with DPP\label{fig:sciq_neighbor_num_random}]{
        \includegraphics[width=0.45\columnwidth]{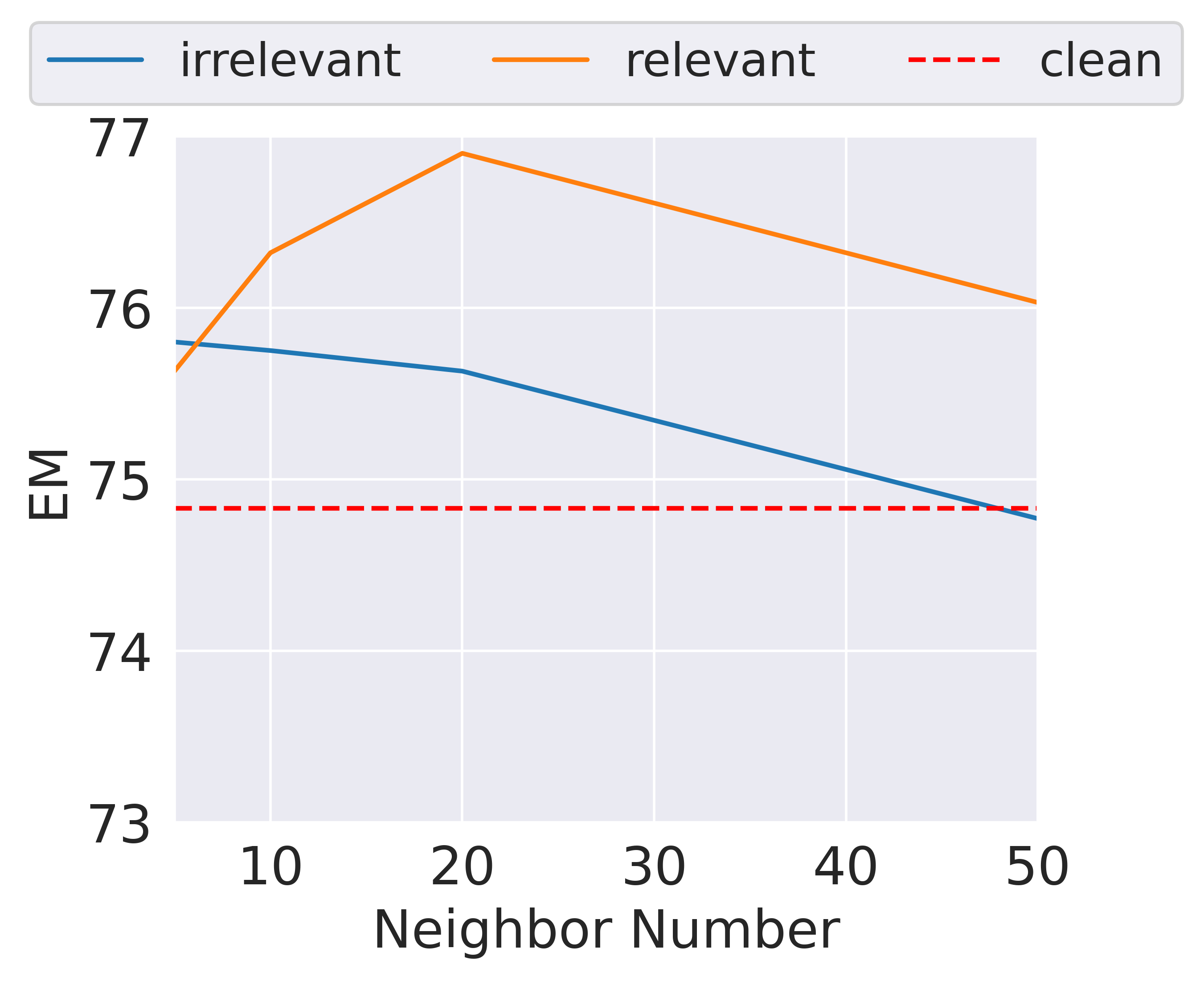}
    }
    \hspace{2mm}
    \subfloat[SCIQ with DPP\label{fig:sciq_neighbor_num_dpp}]{
        \includegraphics[width=0.45\columnwidth]{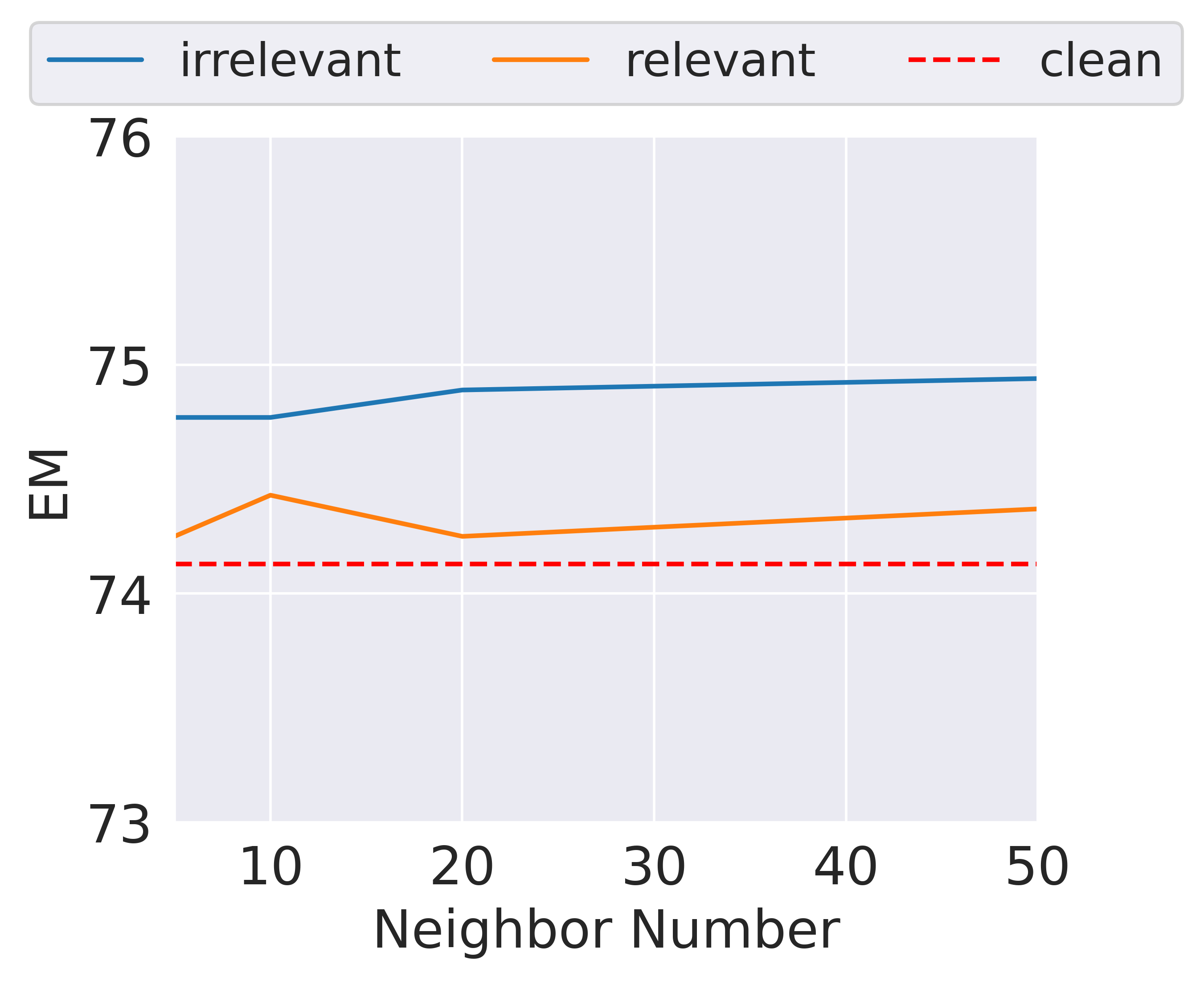}
    }
    \caption{Performance of our method under setting of noise ratio $0.6$ using different number of neighbors.}
    \label{fig:app-more-retriever-neighbor-num}
\end{figure}

As shown in~\autoref{fig:app-more-retriever-neighbor-num}, fewer neighbors only lead to a negligible performance drop on our method, showing the applicability even with a very small number of neighbors to calculate the $\{\mathcal{I}\}_{i=1}^N$.

\subsection{Analysis on $\mathcal{L}(y|x)$}
\label{app:analysis_perplexity}

\begin{figure}[t!]
            \centering
            \includegraphics[width=1.0\columnwidth]{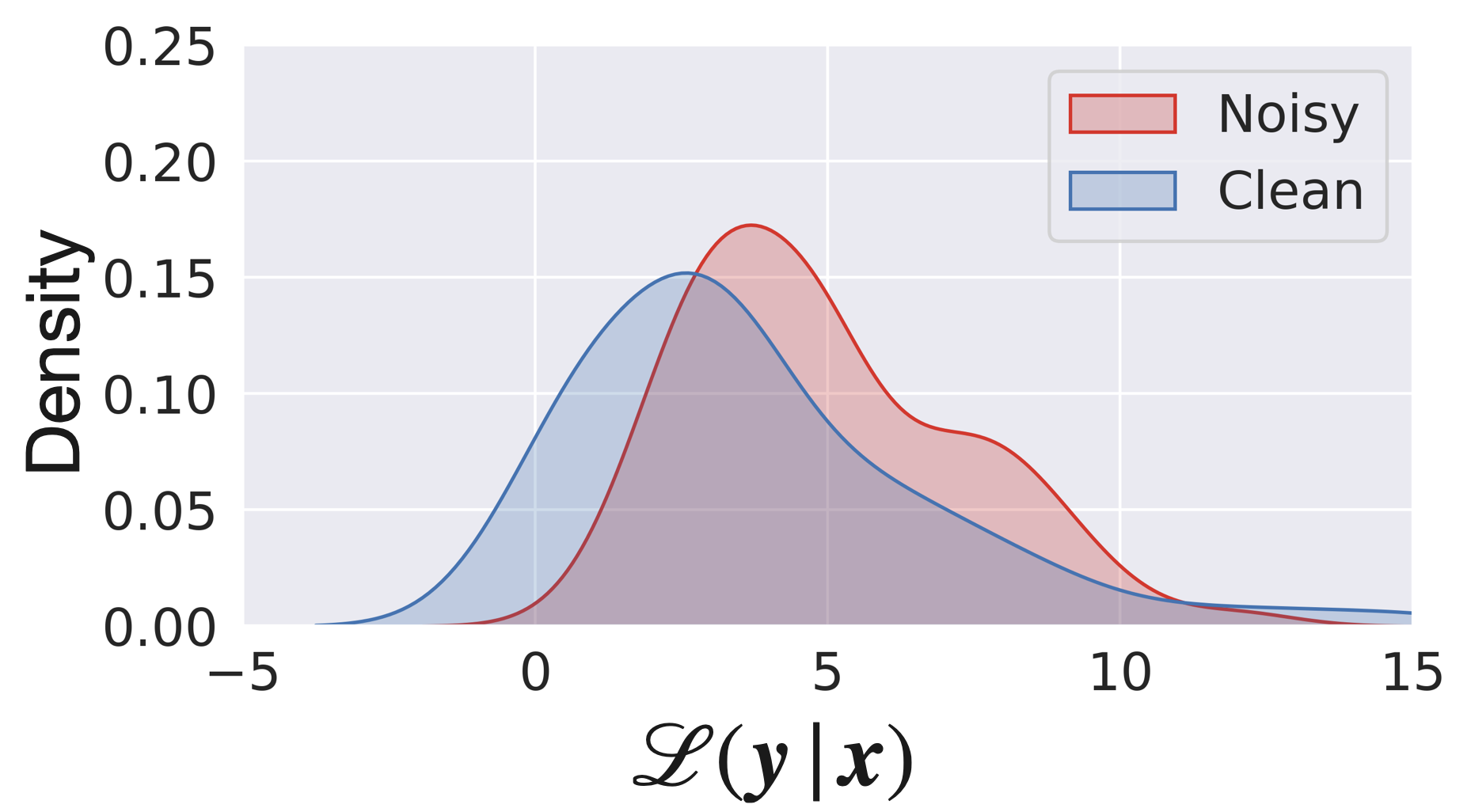}    
            \caption{Distribution of $\mathcal{L}(\boldsymbol{y}|\boldsymbol{x})$ for both noisy and clean samples on NQ. Heavy overlapping between the distributions of noisy and clean samples.}
            \label{fig:ppl-not-work}
\end{figure}

The visualization of $\mathcal{L}(\boldsymbol{y}|\boldsymbol{x})$ on NQ is shown in~\autoref{fig:ppl-not-work}. It shows that using $\mathcal{L}(\boldsymbol{y}|\boldsymbol{x})$ directly is not sufficient to distinguish clean and noisy samples.

\subsection{Robustness on corpus}

\begin{figure}[H]
            \centering
            \includegraphics[width=1.0\columnwidth]{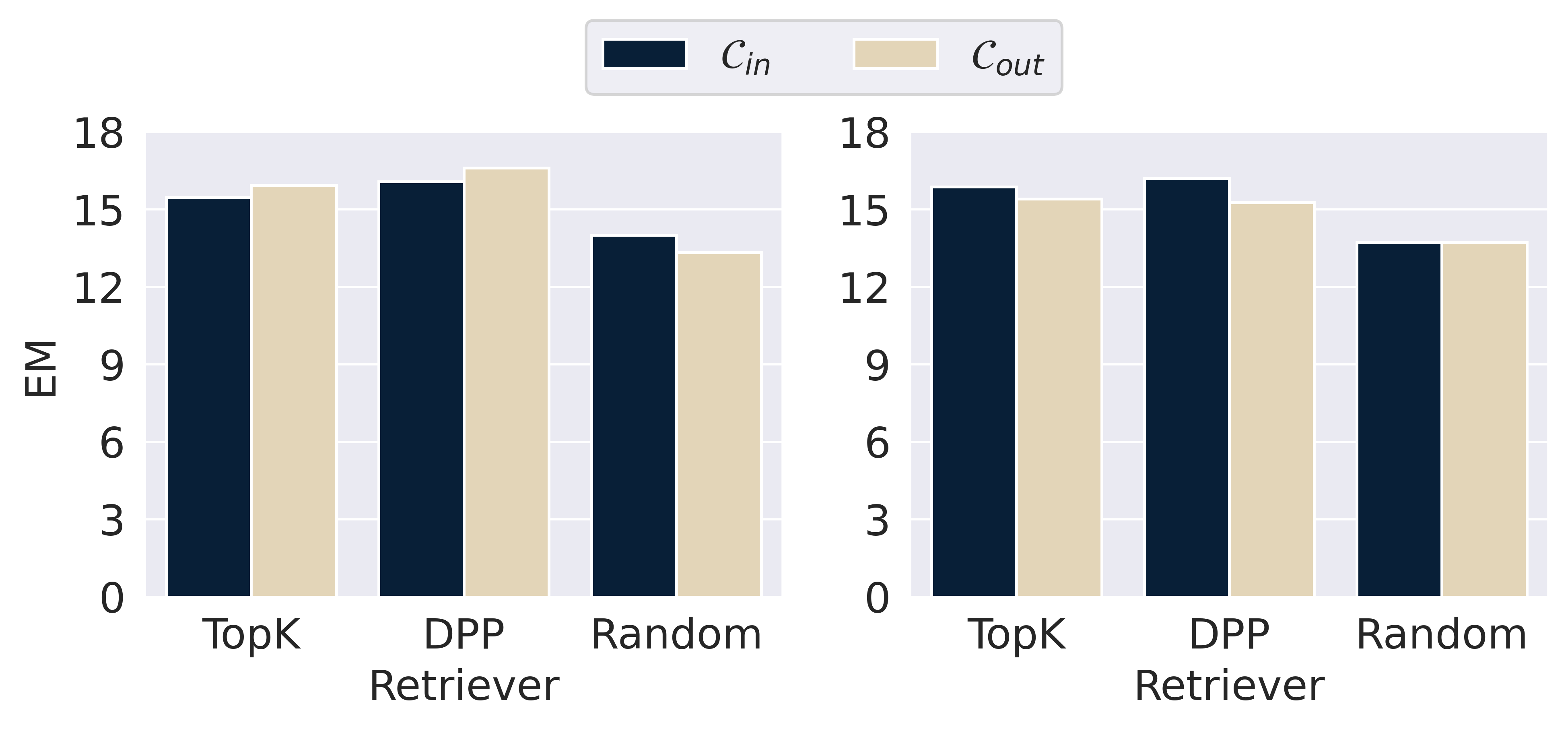}    
            \caption{Comparison between using $\mathcal{C}_{\text{in}}$ and $\mathcal{C}_{\text{out}}$ on NQ under noise ratio $0.6$. Left: irrelevant noise; Right: relevant noise.}
            \label{fig:corpus-nq}
\end{figure}

\begin{figure}[H]
            \centering
        \includegraphics[width=1.0\columnwidth]{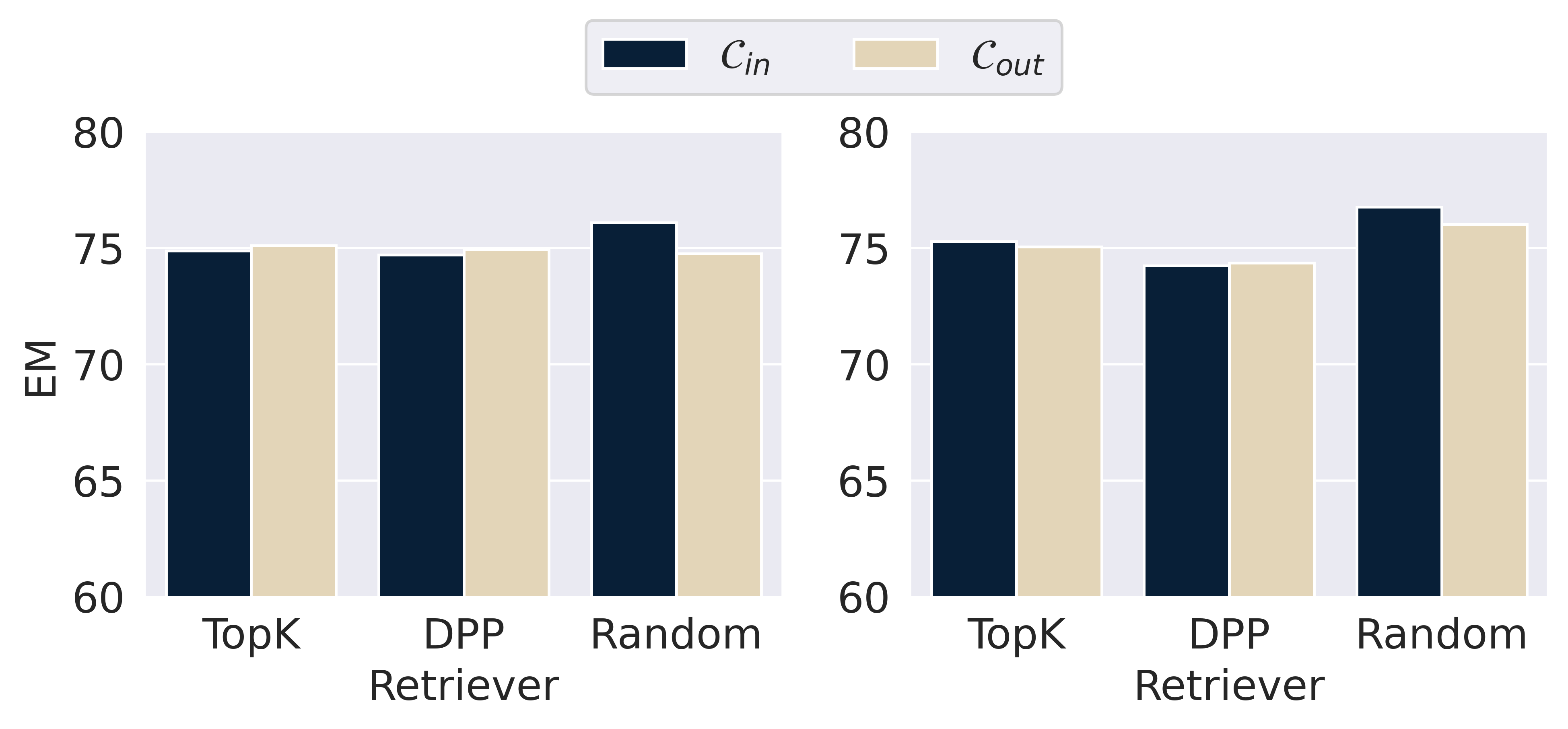}    
            \caption{Comparison between using $\mathcal{C}_{\text{in}}$ and $\mathcal{C}_{\text{out}}$ on SCIQ under noise ratio $0.6$. Left: irrelevant noise; Right: relevant noise.}
            \label{fig:corpus-sciq}
\end{figure}

We conduct the comparison experiment between using $\mathcal{C}_{\text{in}}$ and $\mathcal{C}_{\text{out}}$ under noise ratio $0.6$. \autoref{fig:corpus-nq} presents the result on NQ, and \autoref{fig:corpus-sciq} presents the result on SCIQ.

\subsection{Dependency on choice of LLMs}
\label{app:llm_choice}
We conduct further analysis on how different LLMs benefit from our dual-debiasing approach under $40\%$ relevant noise setting on NQ. And we present the AUC improvement of noisy sample detection using our method $\Delta \text{AUC} = \text{AUC}_{\text{ours}} - \text{AUC}_{\text{naive}}$ in~\autoref{tab:app_llm_auc_delta}, where $\text{AUC}_{\text{naive}}$ is the AUC of using naive per-token loss $\mathcal{L}(\boldsymbol{y} | \boldsymbol{x})$. 
This reveals an important insight: while our method improves performance across all models, smaller LLMs like \code{GPT-Neo-1.3B} demonstrate substantially larger gains. 
This suggests our approach is particularly valuable for resource-constrained scenarios, effectively democratizing robust ICL capabilities across model scales. 
The consistency of improvement across architectures also demonstrates the generalizability of our dual-debiasing framework.
\begin{table}[H]
    \centering
    \resizebox{0.55\linewidth}{!}{
    \begin{tabular}{cc}
    \thickhline
        LLM     & $\Delta \text{AUC} \uparrow$          \\ \hline
        \code{GPT-Neo-1.3B}          & 0.2157  \\
        \code{Gemma-2B}        & 0.1403  \\
        \code{Mistral-7B-v0.1}        & 0.1179         \\
        \code{Llama-2-7B}       & 0.1476         \\
    \thickhline
    \end{tabular}}
    \caption{The improvement of noise detection AUC using our method across various LLMs on NQ with 0.4 relevant noise.}
    \label{tab:app_llm_auc_delta}
\end{table}

\subsection{Results for analysis of failure cases}
\label{app:failure_case}
\begin{figure*}[t]
    \centering
    \includegraphics[width=1\textwidth]{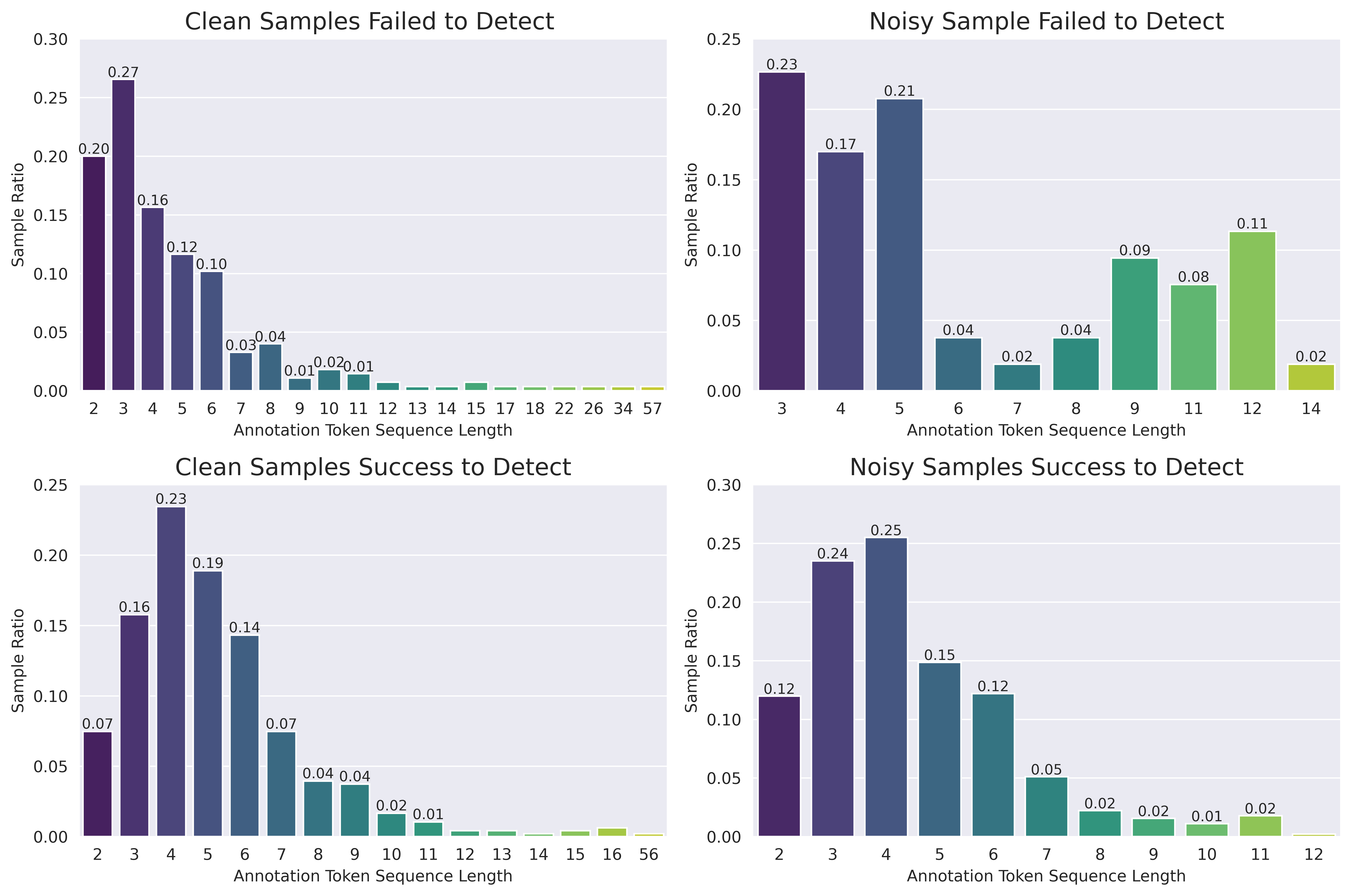}
    \caption{Annotation sequence length distribution for clean/noisy samples that our method failed/successfully detected under WebQ with $40\%$ relevant noise using \code{Llama-2-7B}.}
    \label{fig:app_failed_cases_seq_length}
\end{figure*}

We conducted an in-depth analysis of failure cases on WebQ with $40\%$ relevant noise using \code{Llama-2-7B}.
We present the annotation sequence length distribution for clean/noisy samples that our method failed/successfully detected in~\autoref{fig:app_failed_cases_seq_length}, respectively.

\section{Pseudocode of our method}
\label{app:pseudocode}

We provide the complete pseudocode for our noisy ICL framework in~\autoref{alg:pseudocode}.

\begin{algorithm*}[t]
\caption{Complete pseudocode for proposed dual-debiasing framework for noisy ICL}
\label{alg:pseudocode}
\begin{algorithmic}[1]
\REQUIRE LLM model $\mathcal{M}$, observed training set $\tilde{\mathcal{D}}_{\text{train}}$ with noisy demonstration samples, large corpus $\mathcal{C}$. \\
\textcolor{Skyblue}{{\footnotesize /* Metric Calculation For the Whole Training Set  */ }}
\color{black}
\STATE Initialize empty metric score set $U = \emptyset$ \\
\color{black}
\FOR{each demonstration example $(\boldsymbol{x}_i, \tilde{\boldsymbol{y}}_i) \in \tilde{\mathcal{D}}_{\text{train}}$}
    \STATE Calculate $\mathcal{L}_{\text{de-int}}(\tilde{\boldsymbol{y}}|\boldsymbol{x})$ based on~\autoref{eq:int-debias-definition}
    \STATE Construct neighbors $\mathcal{N}_{\text{DISC}}\left((\boldsymbol{x}_i, \tilde{\boldsymbol{y}}_i)\right)$ using $\mathcal{C}$ as described in~\autoref{sec:neighbor_extrinsic_debias} \\
    \STATE Calculate Sample Cleanliness Score $\mathcal{I}_i$ using $\mathcal{N}_{\text{DISC}}\left((\boldsymbol{x}_i, \tilde{\boldsymbol{y}}_i)\right)$ based on~\autoref{eq:scs_def} and \autoref{eq:extrinsic_bias_estimate} \\
    \STATE Add $\mathcal{I}_i$ to $U$
\ENDFOR \\
\textcolor{Skyblue}{{\footnotesize /* Regular ICL on clean train subset */ }}
\STATE Perform GMM-based noisy sample detection on score set $U=\{I_i\}_{i=1}^N$ as described in~\autoref{sec:noisy_icl_pipeline} \\ 
\STATE Separate training set into clean subset $\tilde{\mathcal{D}}^{\text{clean}}$ and noisy subset $\tilde{\mathcal{D}}^{\text{noisy}}$ \\
\STATE Remove noisy subset, keeping $\tilde{\mathcal{D}}_\text{train}^\prime = \tilde{\mathcal{D}}^{\text{clean}}$ \\
\textcolor{Skyblue}{{\footnotesize /* Metric Calculation For the Whole Training Set  */ }}
\STATE For each test query $\boldsymbol{x}^\text{test}_j \in \mathcal{D}_\text{test}$, perform regular ICL using $\tilde{\mathcal{D}}_{\text{train}}^\prime$ as retrieval pool
\end{algorithmic}
\end{algorithm*}

\section{Generation of neighboring samples}
\label{app:neighbor_generation}

Our neighbor generation process follows a principled approach as described below:  
For each demonstration sample $(\boldsymbol{x}, \tilde{\boldsymbol{y}})$, we construct $N_{\text {neighbor}}$ samples by pairing the original query $\boldsymbol{x}$ with randomly sampled annotations from a given corpus $\mathcal{C}$. Notice that we will drop the annotation sampled from $\mathcal{C}$ if its tokenized sequence length is larger than the neighbor radius $\eta$ defined for $\mathcal{N}_{\text{DISC}}$.
The quality of these neighbors is rigorously controlled through the Edit Distance metric, which quantifies the textual difference between annotations.

We enforce a maximum distance constraint $\eta=\max \left\{T, T_{\text{max}}\right\}$, where $T$ is the observed annotation length and $T_\text{max}$ is the maximum annotation sequence length of the observed annotation from $\tilde{\mathcal{D}}_\text{train}$. 
This formal constraint ensures that neighbors remain within a bounded semantic radius of the original sample, maintaining relevance while providing sufficient diversity for robust debiasing.

\end{document}